\pdfoutput=1

\documentclass[11pt]{article}
\usepackage{enumitem}
\usepackage{booktabs}
\usepackage{float}
\usepackage{color}
\usepackage{acl}
\usepackage{multirow}
\usepackage{times}
\usepackage{latexsym}
\usepackage{subfig}

\usepackage{amsmath}  
\usepackage{amsfonts}  
\usepackage{amssymb}  
\usepackage{hyperref}

\usepackage[T1]{fontenc}

\usepackage[utf8]{inputenc}
\usepackage{xcolor,colortbl}
\usepackage{microtype}

\usepackage{inconsolata}

\usepackage{graphicx}

\usepackage{tcolorbox}

\title{Exposing Numeracy Gaps: A Benchmark to Evaluate Fundamental Numerical Abilities in Large Language Models}

\author{Haoyang LI$^1$, Xuejia CHEN$^1$, Zhanchao XU$^1$, Darian LI$^1$, Nicole HU$^3$,\\ \textbf{Fei TENG$^2$, Yiming LI$^2$\thanks{Corresponding Author}, Luyu QIU$^2$, Chen Jason ZHANG$^1$, Qing LI$^1$, Lei CHEN$^2$} \\
  $^1$Hong Kong Polytechnic University, \\$^2$Hong Kong University of Science and Technology, \\$^3$The Chinese University of Hong Kong\\
  \texttt{haoyang-comp.li@polyu.edu.hk}, 
  \texttt{yliix@connect.ust.hk}
}
 
\begin{document}
\maketitle
\begin{abstract}
Large Language Models (LLMs) have demonstrated impressive capabilities in natural language processing tasks, such as text generation and semantic understanding. However, their performance on numerical reasoning tasks, such as basic arithmetic, numerical retrieval, and magnitude comparison, remains surprisingly poor. This gap arises from their reliance on surface-level statistical patterns rather than understanding numbers as continuous magnitudes. Existing benchmarks primarily focus on either linguistic competence or structured mathematical problem-solving, neglecting fundamental numerical reasoning required in real-world scenarios. To bridge this gap, we propose NumericBench, a comprehensive benchmark to evaluate six fundamental numerical capabilities: number recognition, arithmetic operations, contextual retrieval, comparison, summary, and logic reasoning. NumericBench includes datasets ranging from synthetic number lists to the crawled real-world data, addressing challenges like long contexts, noise, and multi-step reasoning. Extensive experiments on state-of-the-art LLMs, including GPT-4 and DeepSeek, reveal persistent weaknesses in numerical reasoning, highlighting the urgent need to improve numerically-aware language modeling.  
The benchmark is released in:
\url{https://github.com/TreeAI-Lab/NumericBench}.
\end{abstract}

\section{Introduction}
\begin{figure}[t]
	\centering	
 	\vspace{-1em}
	
	\includegraphics[width = 0.45\textwidth]{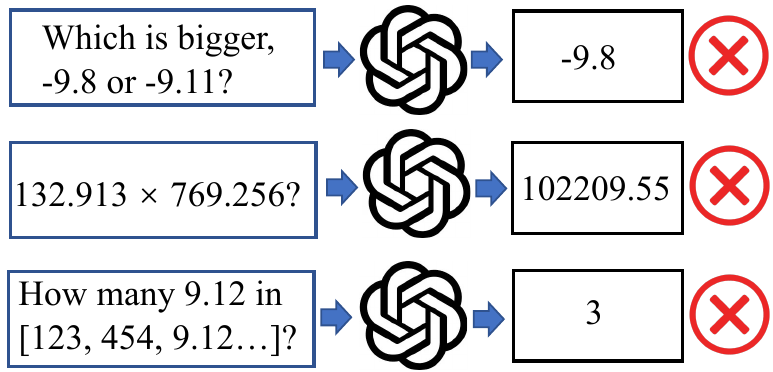}
	\caption{Numerical tasks answered incorrectly by GPT-4o. Details are in Figure~\ref{fig:number_compare},  Figure~\ref{fig:multiplication}, and Figure~\ref{fig:number_couting}.}

	\label{fig:intro_example}
	\vspace{-1em}
\end{figure}

Large language models (LLMs)~\citep{zhao2024surveylargelanguagemodels} 
have demonstrated remarkable capabilities in text generation, semantic understanding, and task adaptation across diverse domains~\citep{ling2024domainspecializationkeymake}. 
Their success is largely attributed to pretraining on vast text corpora using next-token prediction objectives~\citep{he2024lawnexttokenpredictionlarge}, which enables generalization to tasks requiring linguistic creativity, commonsense reasoning, and domain-specific knowledge~\citep{ye-2024-cross}.
However, while LLMs perform well  in text understanding tasks and Olympic mathematics questions~\cite{team2024gemini}, 
they surprisingly struggle with simple numerical tasks such as basic multiplication, comparison, and retrieval. 
As shown in Figure~\ref{fig:intro_example},
 effective GPT-4o fails at simple numerical tasks, even with number comparisons.

Unlike tasks that rely primarily on semantic coherence and linguistic structures, 
numerical reasoning requires a deeper understanding of numbers as continuous magnitudes rather than discrete tokens. 
Current LLMs tend to prioritize surface-level statistical patterns, 
such as lexical co-occurrences and syntactic regularities, which limits their ability to process  numerical operations~\citep{ahn2024largelanguagemodelsmathematical,feng2024numerical,zhou2024transformers}.
As a result, LLMs frequently struggle with tasks involving numeric retrieval, arithmetic operations, 
and magnitude comparisons. These shortcomings highlight an urgent need to systematically evaluate and improve the numerical reasoning capabilities of LLMs.

Current evaluation frameworks for LLMs prioritize either linguistic competence or formal mathematical problem-solving. 
For instance, 
semantic-oriented benchmarks~\citep{vulic-etal-2020-probing}, such as GLUE~\citep{wang2018glue}, 
SuperGLUE~\citep{wang2019superglue}, 
and SimpleQA~\citep{wei2024measuring}, 
primarily assess linguistic competence and semantic understanding, 
while math-oriented benchmarks, such as MathQA~\citep{amini2019mathqa}, 
GSM8K~\citep{cobbe2021training}, 
and MathBench~\citep{liu2024mathbench},
focus on structured algebraic or geometric tasks. 
However, these approaches neglect the basic demands of real-world numerical reasoning applications and fundamental numerical abilities, 
where numbers are often embedded in unstructured and noisy context data. 
For example, analyzing fluctuating stock prices or weather   requires basic  numeric retrieval, comparison, and summery abilities.

To address the limitations of existing benchmarks, we propose  a comprehensive benchmark  NumericBench, 
which consists of six general datasets, i.e., arithmetic numbers, mixed-number-strings, number lists, stock, weather, and numerical sequences with patterns.
Unlike prior benchmarks, NumericBench 
systematically evaluate the six fundamental numerical abilities of LLMs:
\textbf{(1) Number Recognition:} 
It evaluates the ability of LLMs to identify numbers within dense strings.
\textbf{(2) Arithmetic Operation:} 
It tests basic arithmetic operations, including addition, subtraction, multiplication, and division.
\textbf{(3) Contextual Retrieval:}
It evaluates LLMs to retrieve specific numerical values from a given context or a number list.
\textbf{(4) Comparison:}
It determine relationships between values, such as comparing price differences.
\textbf{(5) Summary:}
It target to summarize trends (e.g., determining the number of consecutive days a stock price increases) and data aggregation.
\textbf{(6) Logic Reasoning:}
It tests whether LLMs can understand numerical patterns and predict the next value, which is widely used in time-series forecasting, such as weather and traffic prediction.

By integrating six datasets, ranging from synthetic numerical lists to crawled real-world data, NumericBench evaluates six fundamental numerical abilities of LLMs across various scenarios, such as noisy contexts.
Our experimental analysis over various effective series of LLMs, including 
GPT-4~\citep{achiam2023gpt}, 
DeepSeek~\citep{liu2024deepseek}, 
and Llama~\citep{touvron2023llama}, 
reveals persistent weaknesses in handling these fundamental numerical tasks. 
We further analyze five potential reasons behind these numerical reasoning gaps, including tokenizer, training corpora, training paradigms, positional embeddings, and architectural constraints.
These findings offer actionable insights to guide future  numerical ability improvements for LLMs. 
Given that numerical reasoning is critical for real-world problem-solving, it represents a cornerstone for the development of Artificial General Intelligence (AGI). 
This emphasizes the urgent need to advance numerically-aware language modeling.
The contributes of this paper are summarized as follows:
\begin{itemize}[leftmargin=*]
	\item 
	We propose a comprehensive NumericBench, which integrates diverse datasets and reflects real-world challenges, such as handling noisy or domain-specific data (e.g., stock and weather).
	NumericBench evaluates six fundamental numerical reasoning abilities, including number recognition, arithmetic operations, 
	contextual retrieval, comparison, summary, and logic reasoning.

	\item  Extensive experiments on effective LLMs (e.g., GPT-4, DeepSeek, Llama) reveal persistent weaknesses in numerical reasoning tasks, including basic arithmetic, comparison, and logic reasoning. This highlights the need for more numerically-aware modeling approaches.

	\item We identify key challenges behind the numerical reasoning gaps in LLMs, such as tokenization practices, training paradigms, positional embeddings, and architectural constraints. These insights provide actionable guidance for future model improvements.
	
\end{itemize}

\section{Preliminary and Related Works}
In this section, we first introduce large language models and then present existing benchmarks.

\subsection{Large Language Models}
Large language models (LLMs), such as GPT-4~\citep{achiam2023gpt}, 
DeepSeek~\citep{liu2024deepseek}, 
PaLM~\citep{anil2023palm}, 
and Llama~\citep{touvron2023llama}, 
have revolutionized natural language processing (NLP) through their ability to generate coherent text~\citep{cho2019coherentcohesivelongformtext}, 
answer questions~\citep{chen2024analyze}, 
and adapt to diverse tasks~\citep{wang2025graph,jiang2024survey}. 
Their success stems from pretraining on vast text corpora using next-token prediction objectives, 
which enable generalization on tasks requiring semantic understanding, commonsense reasoning, and linguistic creativity.
However, this training paradigm encourages LLMs to prioritize surface-level statistical patterns 
(e.g., lexical co-occurrences, syntactic regularities) rather than numerically grounded reasoning~\citep{bachmann2024pitfalls}. 
Consequently, LLMs treat numbers as discrete tokens rather than continuous magnitudes, inherently limiting their ability to understand exact numerical semantics. 
This leads to errors in numeric retrieval, arithmetic operations, and magnitude comparisons~\citep{qiu2024dissecting}.

\subsection{Benchmarks on Large Language Models}

Existing benchmarks~\cite{li2024survey,chang2024survey,zhao2023survey} for evaluating LLMs primarily fall into two categories, 
i.e., \textit{semantic-oriented} and \textit{math-oriented} benchmarks.
Specifically,
\textit{semantic-oriented} 
benchmarks, 
such as GLUE~\citep{wang2018glue},  
SuperGLUE~\citep{wang2019superglue}, 
SimpleQA~\citep{wei2024measuring}, 
and LongBench~\citep{bai2023longbench},
focus on semantic understanding and linguistic competence, 
testing skills like textual entailment, 
commonsense reasoning, and domain-specific knowledge (e.g., science and law). 
While effective for assessing linguistic proficiency, these benchmarks largely overlook numerical reasoning.
On the other hand,
\textit{math-oriented} benchmarks~\cite{gao2025gllava,li2024forewarned,cobbe2021training}, 
such as MathQA~\citep{amini2019mathqa}, 
GSM8K~\citep{cobbe2021training}, 
and MathBench~\citep{liu2024mathbench},
target mathematical problem-solving (e.g., algebra, calculus) or extractive question-answering with numerical answers. 
However, these datasets emphasize well-formed mathematical problems  in controlled and clean settings.
Consequently,
\textit{math-oriented} benchmarks poorly evaluate numerical retrieval and reasoning in real-world conditions, where noise, and contextual complexity (e.g., multi-step financial workflows or long stock sequences) are common.


Considering that numeric retrieval and reasoning are critical for real-world applications~\cite{yang2025numbercookbook}, 
such as finance~\citep{islam2023financebench} and weather forecasting~\cite{zhang2024self}, 
we propose \textbf{\textit{NumericBench}} to systematically evaluate the fundamental numerical abilities of  LLMs on intensive tasks, such as precise value retrieval, dynamic comparisons, and arithmetic-logic reasoning.

\section{NumericBench}
In this section, we present our created  NumericBench, which is specifically designed to evaluate fundamental numerical capabilities of LLMs. 
NumericBench consists of diverse datasets and tasks, 
enabling a systematic and comprehensive evaluation.
We discuss the datasets included in NumericBench, the key abilities it evaluates, and the methodology for benchmark generation.

\begin{table*}[t]
	\caption{NumericBench statistics. R: contextual retrieval, C: comparison, S: summary, L: logic reasoning. The token count is calculated based on tiktoken, which is the tokenizer used by Llama3~\cite{grattafiori2024llama3herdmodels}. The sentences used for token calculation include both the context and the question.}
	\centering
	\renewcommand{\arraystretch}{1.15} 
	\setlength{\tabcolsep}{1.5pt} 
\resizebox{\textwidth}{!}{
	\begin{tabular}{c|c|c|c|c}
		\toprule
		\textbf{Data} & \textbf{Format} & \textbf{Questions} & \textbf{\# Instance} & \textbf{Avg Token} \\ \midrule
		
		\multirow{3}{*}{} 
		& \multirow{3}{*}{} 
		& \begin{tabular}[c]{@{}c@{}}R: What is the index of the first occurrence\\ of the number -3095 in the list?\end{tabular} 
		& 500 & 3704.23 \\ \cline{3-5}
		
		\textbf{\begin{tabular}[c]{@{}c@{}}Number\\ List\end{tabular}}
		& $[69, -1, 6.1, \ldots, 5.7]$
		& \begin{tabular}[c]{@{}c@{}}C: Which index holds the smallest number\\
			 in the list between the indices 20 and 80?\end{tabular} 
		& 500 & 3685.57  \\ \cline{3-5}
		
		& & \begin{tabular}[c]{@{}c@{}}S: What is the average of the index of\\
			 top 30 largest numbers in the list?\end{tabular} 
		& 500 & 3654.78 \\ \midrule
		
		\multirow{3}{*}{} 
		& \multirow{3}{*}{
		\begin{tabular}[c]{@{}c@{}}
			\{date: 2024-06-19,\\
			close\_price: 9.79, \\
			open\_price: 9.4, \\
			\ldots \\
			PE\_ratio: 4.5416\}
		\end{tabular}
		} 
		& \begin{tabular}[c]{@{}c@{}}
			R: On which date did the close price\\
			 of stock firstly reach 61.76 yuan?
		\end{tabular}
		& 500 & 27585.35 \\ \cline{3-5}
		
		\textbf{Stock}
		& 
		& \begin{tabular}[c]{@{}c@{}}
			C: Among the top-45 trading value days, which\\
			 date did the stock have the lowest close price?
		\end{tabular}
		 & 500 & 27595.40 \\ \cline{3-5}
		
		& & \begin{tabular}[c]{@{}c@{}} 
			S: How many days had the close price higher than\\
			 the open price from 2024-07-31 to 2024-12-13?
		\end{tabular}	
		& 500 & 27561.29 \\ \midrule
		
		\multirow{3}{*}{} 
		& \multirow{3}{*}{
		\begin{tabular}[c]{@{}c@{}}
			\{date: 2024-07-21,\\
			pressure\_msl: 999.96,\\
			dew\_point\_2m: 26.25,\\
			\ldots \\
			cloud\_cover: 61.5\}
		\end{tabular}
		} 
		& \begin{tabular}[c]{@{}c@{}} 
			R: On which date did the dew point temperature\\
			 at two meters lastly drop below 9.15°C?
		\end{tabular}
		& 500 & 27359.26 \\ \cline{3-5}
		
		\textbf{Weather}
		& & \begin{tabular}[c]{@{}c@{}} 
			C: On which date did the MSL pressure reach its\\
			highest value when the cloud cover was below 9\%?
		\end{tabular}
		& 500 & 27368.19 \\ \cline{3-5}
		
		& & \begin{tabular}[c]{@{}c@{}} 
			S: What was the average temperature at two meters\\
			when the relative humidity exceeded 78.56\%?
		\end{tabular}
		& 500 & 27331.21 \\ \midrule
		
		\textbf{Sequence} 
		& $[0.34, 3, 6, \ldots, 111]$ 
		& L: What is the next number in the sequence? & 500 & 677.57 \\ \midrule
		
		\textbf{\begin{tabular}[c]{@{}c@{}}Arithmetic \\Operation\end{tabular}} 
		& \begin{tabular}[c]{@{}c@{}} 
		$a: 6.755,
		b: -1.225$
		\end{tabular}
		& \begin{tabular}[c]{@{}c@{}} 
		 $Q_{oper}$: What is the result of $a + b$?\\
		 $Q_{context}$: What is the result of $a $ plus $b$?
		 
		\end{tabular}
		& 12000 & 112.00 \\ \midrule
		
		\textbf{\begin{tabular}[c]{@{}c@{}}Mixed-number-string\\ Sequence\end{tabular}} 
		& \begin{tabular}[c]{@{}c@{}} 
		$effV2\ldots x98o7Lo$
		\end{tabular}
		& \begin{tabular}[c]{@{}c@{}} 
		How many numbers are there in the string? Note\\
		that a sequence like 'a243b' counts as a single number.
		\end{tabular}
		& 2000 & 196.53 \\ \bottomrule

	\end{tabular}
}
	\label{tab:data_stat}
	
\end{table*}

\subsection{Numeric Dataset Collection}
NumericBench offers a diverse collection of numerical datasets and questions designed to reflect real-world scenarios. 
This variety ensures that LLMs are thoroughly tested on their fundamental  abilities on numerical data.

\noindent\textbf{Number List Dataset.}
The synthetic number list dataset consists of randomly generated numerical values, including both integers and floating-point numbers.
presented as ordered or unordered lists.
Numbers in lists are one of the simplest and most fundamental data representations encountered in real-world scenarios.
Despite their simplicity, retrieving, indexing, comparison, and summary on numbers can verify the fundamental numerical ability of LLMs. 
This dataset serves as a fundamental dataset of how well LLMs understand numerical values as discrete entities.

\noindent\textbf{Stock Dataset.}
The time-series  stock dataset is crawled from Eastmoney website~\cite{eastmoney}, 
which has eighteen attributes, such as stock close prices, open prices,  trading volumes, and price-earnings ratios, over time.
Stock  data reflects dynamic, real-world numerical reasoning challenges that involve trend analysis, comparison, and decision-making under uncertainty,  representing real-world financial workflows.

\noindent\textbf{Weather Dataset.}
The weather dataset is crawled from Open-Meteo python API~\citep{openmeteo}, which includes data related to weather metrics, such as temperature, precipitation, humidity, and wind speed. 
The data is presented across various longitude and latitude.

\noindent\textbf{Numeric Sequence  Dataset.}
The synthetic numeric sequence dataset comprises sequences of numbers generated by arithmetic or geometric sequence with various patterns, such as Fibonacci Sequence. 
Tasks require identifying patterns, predicting the next number, or reasoning about relationships between numbers.
Numerical sequences test the logic reasoning capabilities of LLMs, requiring pattern recognition and multi-step reasoning. We introduce structured challenges that are common in mathematics and algorithmic reasoning.

\noindent\textbf{Arithmetic Operation Dataset.}
The dataset comprises 12,000 pairs of simple numbers, each undergoing addition, subtraction, multiplication, and division operations. Each pair of numbers, $a$ and $b$, consists of $k$-digit integers with three decimal places, where $k \in \{1, 2, \cdots, 6\}$. 
For each value of $k$, there are 2,000 pairs, evenly distributed across the four basic operations (i.e, $+, -,  *, /$), with 500 pairs per operation. 
This dataset is to evaluate the fundamental mathematical operation capabilities of LLMs, simulating the majority of mathematical calculation requirements in real-world scenarios.

\noindent\textbf{Mixed-number-string Sequence Dataset.}
The dataset consists of alphanumeric strings of varying lengths $\{50, 100, 150, 200\}$, each containing a randomized mix of letters and digits. For each string length, 500 samples are generated, resulting in a total of 2,000 samples. Each sample includes a query asking for the count of contiguous numeric sequences within the string, where sequences like "a243b" count as a single number. This dataset is designed to assess the ability of LLMs to identify and count numeric sequences.

\subsection{Fundamental Numerical Ability}
NumericBench is designed to comprehensively evaluate six fundamental numerical reasoning abilities of LLMs, which are essential for solving real-world numeric-related tasks.


\noindent\textbf{Contextual Retrieval Ability.}
Contextual retrieval ability evaluates how well LLMs can locate, extract, and identify specific numerical values or their positions within structured or unstructured data. 
This includes tasks like finding a specific number in a list, retrieving values , and indexing numbers based on their order.
For example, as shown in Table~\ref{tab:data_stat}, it evaluates LLMs on tasks such as retrieving stock prices and identifying key values within numerical lists or domain-specific data (e.g., stock market and weather-related information).
This ability is fundamental to numerical reasoning because it forms the foundation for higher-order tasks, such as comparison, aggregation, and logic reasoning.

\noindent\textbf{Comparison Ability.}
Comparison ability evaluates how well LLMs can compare numerical values to determine relationships such as greater than, less than, or equal to, and identify trends or differences in datasets. 
Comparison is vital for logic reasoning and decision-making, as many real-world tasks depend on accurate numerical evaluation. 
For instance,  as shown in Table~\ref{tab:data_stat},   comparing prices is essential in stock  for assessing performance, while weather forecasting requires analysis of temperature or precipitation trends over time.

\noindent\textbf{Summary Ability.}
Summary ability assesses the LLM’s capacity to aggregate numerical data into concise insights, such as calculating totals, averages, or other statistical metrics. 
Summarization is critical for condensing large datasets into actionable information, enabling decision-making based on aggregated insights rather than raw data. 
This ability is indispensable in domains like electricity usage analysis, where summarizing hourly or daily consumption helps forecast bills, in business reporting for aggregating sales and revenue data to evaluate performance, 
and in healthcare analytics to monitor trends in patient metrics over time.

\noindent\textbf{Logic Reasoning Ability.}
Logic Reasoning Ability measures the LLM’s ability to perform multi-step operations involving numerical data, 
such as recognizing patterns, inferring rules, and applying arithmetic or geometric reasoning to solve complex problems. Logic reasoning extends beyond simple numerical tasks and reflects the LLM’s capacity for deeper, structured thinking. 
This ability is crucial for algorithm design, where solving problems involving numeric sequences or patterns is essential. It is also required in scientific research for identifying relationships and correlations in data.

\noindent\textbf{Arithmetic Operation Ability.}
It reflects the LLM's capacity to perform mathematical calculations accurately. Such ability is essential for tasks involving numerical computations, such as  automated machine learning through LLMs.

\noindent\textbf{Number Recognition  Ability.}
This measures the LLM's proficiency in identifying and interpreting numerical information within a given context. It represents a fundamental requirement for handling numeric-based tasks effectively.

\subsection{NumericBench Generation}
We use the number list, stock, and weather datasets to evaluate the contextual retrieval, comparison, and summary abilities of LLMs. 
Specifically, for each ability and each dataset, we prepare a set of questions designed to assess the corresponding target ability.
As shown in Table~\ref{appx:number_question}, Table~\ref{appx:stock_question}, and Table~\ref{appx:weather_question} in Appendix, there are nine question sets in total, covering three abilities across three datasets. 
When evaluating a specific ability (e.g., contextual retrieval) on a specific dataset (e.g., stock data), we randomly select one question from the corresponding question set for each data instance (e.g., a stock instance) 
and manually label the answer. This approach enables us to generate question-answer pairs for each ability on the number list, stock, and weather datasets.

Moreover, we generate a logic reasoning dataset with 500 long sequences using general term formulas. The target value of each sequence is removed to form an inference task, as detailed in Table~\ref{tab:data_stat} of our paper.
For arithmetic operations and number counting in the strings dataset, the question format is straightforward, as illustrated in Table~\ref{tab:data_stat}. These questions  evaluate the basic arithmetic operation and number recognition abilities of LLMs.

\section{Experiments}

\subsection{Experiment Setting}

\noindent\textbf{Benchmarks and Evaluated Protocols.}
The statistic of  NumericBench is provided in Table~\ref{tab:data_stat}.
Also, we set the exact answer for mixed-number-string dataset, set the computed answer to two decimal places for arithmetic datasets, and  set the answer of each question as a single choice (e.g., A, B, or C) for other datasets to reliably evaluate LLMs~\citep{bai2024longbench}.
The evaluation metric is accuracy.

\begin{table*}[t]
	\centering
		\vspace{-2em}
	\setlength\tabcolsep{2pt}
	\footnotesize
	\caption{Evaluation of LLMs on numerical contextual retrieval, comparison, and summary tasks across number list, stock, and weather datasets. 
		Also, * indicates that scores are calculated based on a short subset of outputs, as these models cannot handle  long contexts and exhibit disruption when tested on longer instances.}
	\begin{tabular}{c|ccc|ccc|ccc|c}
		\toprule
		\multirow{2}{*}{\textbf{Model}} & \multicolumn{3}{c|}{\textbf{Retrieval}}                                                         & \multicolumn{3}{c|}{\textbf{Comparison}}                                                           & \multicolumn{3}{c|}{\textbf{Summary}}                                                           & \textbf{Logic}    \\ \cmidrule{2-11} 
		
		& \multicolumn{1}{c}{\textbf{Number}} & \multicolumn{1}{c}{\textbf{Stock}} & \textbf{Weather} & \multicolumn{1}{c}{\textbf{Number}} & \multicolumn{1}{c}{\textbf{Stock}} & \textbf{Weather} & \multicolumn{1}{c}{\textbf{Number}} & \multicolumn{1}{c}{\textbf{Stock}} & \textbf{Weather} & \textbf{Sequence} \\ \midrule
		
		\textbf{\texttt{Random}} & \multicolumn{1}{c}{12.5}                  & \multicolumn{1}{c}{12.5}               &          12.5        & \multicolumn{1}{c}{12.5}                  & \multicolumn{1}{c}{12.5}               &                 12.5 & \multicolumn{1}{c}{12.5}                  & \multicolumn{1}{c}{12.5}               &      12.5            &         12.5          \\ \midrule

		\textbf{\texttt{Llama-3.1-8B-Inst}}& \multicolumn{1}{c}{22.8}                  & \multicolumn{1}{c}{14.4}               &      12.5          & \multicolumn{1}{c}{19.5}                  & \multicolumn{1}{c}{11.7}               &     13.7            & \multicolumn{1}{c}{18.1}                  & \multicolumn{1}{c}{13.8}               &       13.9*          &       18.2            \\  
		
		\textbf{\texttt{Llama-3.1-70B-Inst}}& \multicolumn{1}{c}{37.3}                  & \multicolumn{1}{c}{17.4}               &     33.0             & \multicolumn{1}{c}{28.3}                  & \multicolumn{1}{c}{15.0}               &     28.7             & \multicolumn{1}{c}{24.7}                  & \multicolumn{1}{c}{16.4}               &       15.2           &     17.8              \\  
		
		\textbf{\texttt{Llama-3.3-70B-Inst}}& \multicolumn{1}{c}{44.4}                  & \multicolumn{1}{c}{19.4}               &      32.6            & \multicolumn{1}{c}{31.5}                  & \multicolumn{1}{c}{13.8}               &       35.8           & \multicolumn{1}{c}{26.3}                  & \multicolumn{1}{c}{16.8}               &   18.0               &     18.6              \\  
		
		\textbf{\texttt{Llama-3.1-405B-Inst}}& \multicolumn{1}{c}{44.6}                  & \multicolumn{1}{c}{26.8}               &          23.8        & \multicolumn{1}{c}{25.1}                  & \multicolumn{1}{c}{14.8}               &     29.8             & \multicolumn{1}{c}{32.9}                  & \multicolumn{1}{c}{17.0}               &    16.1              &     16.6              \\  
		
		\textbf{\texttt{Llama-3.1-Nemotron-70B-Inst}}& \multicolumn{1}{c}{41.6}                  & \multicolumn{1}{c}{19.3}               &        33.5          & \multicolumn{1}{c}{26.6}                  & \multicolumn{1}{c}{13.7}               &      33.6            & \multicolumn{1}{c}{29.4}                  & \multicolumn{1}{c}{16.5}               &     17.0             &     16.4              \\  
		
		\textbf{\texttt{Qwen2.5-7B-Inst}}& \multicolumn{1}{c}{20.2}                  & \multicolumn{1}{c}{17.3}               &    22.8              & \multicolumn{1}{c}{24.8}                  & \multicolumn{1}{c}{17.8}               &      18.8            & \multicolumn{1}{c}{18.5}                  & \multicolumn{1}{c}{11.7}               &     13.8             &    14.4               \\  
		\textbf{\texttt{Qwen2.5-72B-Inst}}& \multicolumn{1}{c}{28.8}                  & \multicolumn{1}{c}{41.4*}               &       14.0*           & \multicolumn{1}{c}{28.0}                  & \multicolumn{1}{c}{26.0*}               &       31.0*           & \multicolumn{1}{c}{31.9}                  & \multicolumn{1}{c}{18.8*}               &        16.4*          &      19.0             \\  
		\textbf{\texttt{GLM-4-Long}}& \multicolumn{1}{c}{26.5}                  & \multicolumn{1}{c}{19.5}               &       19.4           & \multicolumn{1}{c}{18.9}                  & \multicolumn{1}{c}{14.8}               &      21.6            & \multicolumn{1}{c}{20.8}                  & \multicolumn{1}{c}{10.8 }               &      10.5            &        17.6           \\  
		
				\textbf{\texttt{Deepseek-V3}}& \multicolumn{1}{c}{47.2}                  & \multicolumn{1}{c}{47.5}               &       27.6          & \multicolumn{1}{c}{27.0}                  & \multicolumn{1}{c}{22.5}               &       35.8          & \multicolumn{1}{c}{21.8}                  & \multicolumn{1}{c}{13.0}               &       15.1          &   15.8                \\  
		
		\textbf{\texttt{GPT-4o}}& \multicolumn{1}{c}{41.7}                  & \multicolumn{1}{c}{37.5}               &        48.8          & \multicolumn{1}{c}{30.6}                  & \multicolumn{1}{c}{33.0}               &       64.2           & \multicolumn{1}{c}{11.6}                  & \multicolumn{1}{c}{17.4}               &      16.5            &        14.6           \\ 
		
		\textbf{\texttt{\textcolor{black}{o3-mini}}} & \textcolor{black}{96.8} & \textcolor{black}{68.6} & \textcolor{black}{78.1} & \textcolor{black}{84.5} & \textcolor{black}{69.9} & \textcolor{black}{83.9} & \textcolor{black}{68.6} & \textcolor{black}{68.5} & \textcolor{black}{72.8} & \textcolor{black}{\textbf{66.4}} \\

		\textbf{\texttt{\textcolor{black}{Deepseek-R1}}} & \textcolor{black}{73.6} & \textcolor{black}{81.3} & \textcolor{black}{93.6} & \textcolor{black}{64.5} & \textcolor{black}{85.3} & \textcolor{black}{98.8} & \textcolor{black}{62.1} & \textcolor{black}{64.3} & \textcolor{black}{61.7} & \textcolor{black}{65.4} \\

		\midrule
		
		\textbf{\texttt{Human Evaluation}}& \multicolumn{1}{c}{\textbf{100}}                  &  \multicolumn{1}{c}{\textbf{100}}               &      \textbf{100}            & \multicolumn{1}{c}{\textbf{100}}                  & \multicolumn{1}{c}{\textbf{100}}               &        \textbf{100}          & \multicolumn{1}{c}{\textbf{100}}                  & \multicolumn{1}{c}{\textbf{100}}               &          \textbf{100}        &               52.6   \\ \bottomrule
	\end{tabular}
	\label{tab:main_experiments}
\end{table*}
\begin{figure*}[t]
		\vspace{-1em}
	\centering 	
	\subfloat[Contextual  Retrieval]	
	{\centering\includegraphics[width=0.33\linewidth]{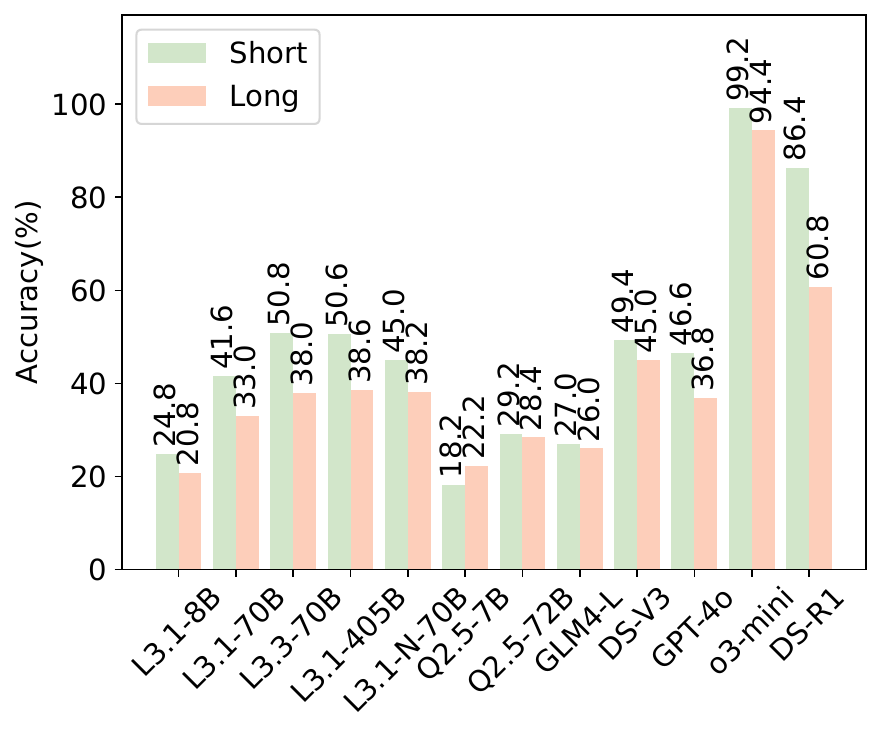}}
	\hfill
	\subfloat[Comparison]
	{\centering\includegraphics[width=0.33\linewidth]{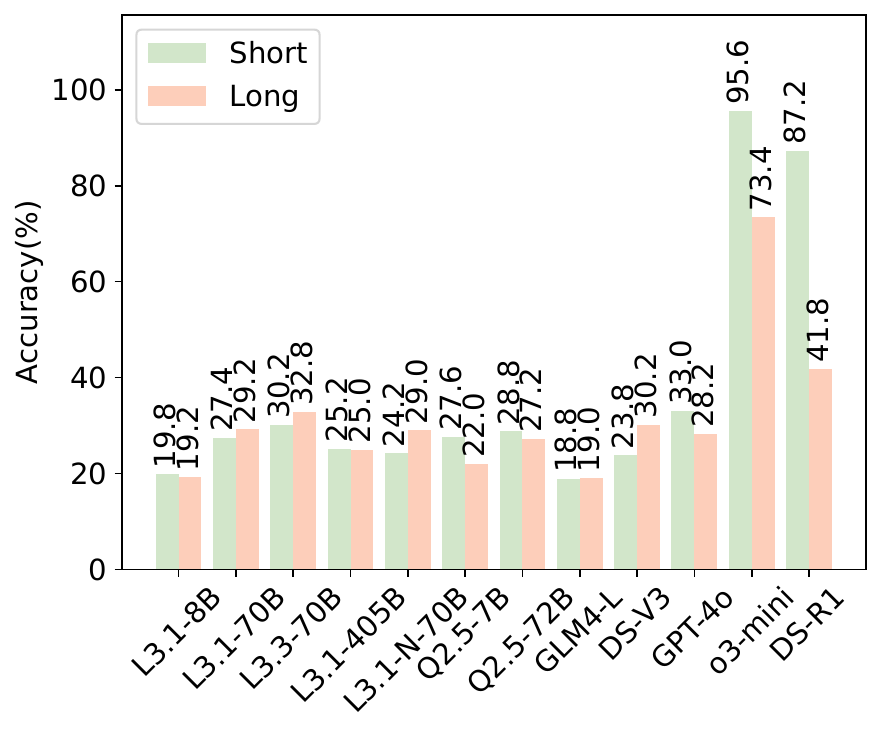}}
	\subfloat[Summary]	
	{\centering\includegraphics[width=0.33\linewidth]{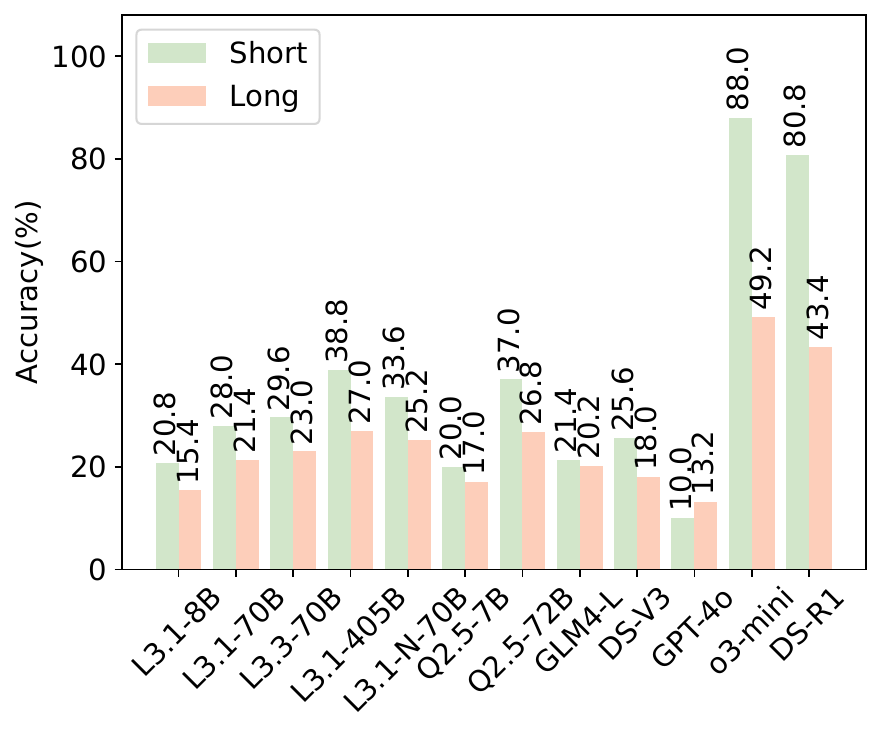}}
	\hfill
	\caption{\textcolor{black}{Evaluation on short and long context on number list.}}
	\label{fig:length_number}
	
\end{figure*}
\noindent\textbf{Evaluated Models.}
To comprehensively evaluate the retrieval and reasoning abilities of state-of-the-art and widely-used LLMs on numeric data, 
we benchmark over 10 popular LLMs with our constructed NumericBench, as follows.
\begin{itemize}[leftmargin=*]
	\item \textbf{The Llama Series~\citep{grattafiori2024llama3herdmodels}.} include Llama-3.1-8B-Instruct, Llama-3.1-70B-Instruct, Llama-3.1-405B-Instruct, 
	Llama-3.3-70B-Instruct and Llama-3.1-Nemotron-70B-Instruct.
	\item \textbf{The Qwen Series~\citep{qwen2025qwen25technicalreport}.} include the effective Qwen2.5-7B-Instruct and Qwen2.5-72B-Instruct. 
	\item \textbf{The GLM Series~\citep{glm2024chatglm}.} We use GLM4-Long to run the benchmark, since it is the commonly used in GLM series.
	
	\item \textbf{The Deepseek Series~\citep{liu2024deepseek,deepseekai2025deepseekr1incentivizingreasoningcapability}.} We use Deepseek-V3 and Deepseek-R1 to run the benchmark.
	\item  \textbf{The GPT Series~\cite{achiam2023gpt}.} We evaluate GPT-4o and OpenAI o3-mini on our proposed benchmark. 
\end{itemize}

We attempted to conduct experiments   on various math-oriented LLMs, such as Metamath-Llemma-7B~\citep{yu2023metamath}, Deepseek-Math-7B-instruct~\citep{deepseek-math}, InternLM2-Math-7B~\citep{ying2024internlmmathopenmathlarge} and MAmmoTH-7B~\citep{yue2023mammoth}.
However, these models fail during experiments for various reasons such as overly long output sequence length and limited input sequence length. Fail cases are demonstrated in the Figure~\ref{fig:fail_internlm},  Figure~\ref{fig:fail_ds_math},  Figure~\ref{fig:fail_llemma}, and  Figure~\ref{fig:fail_mammoth} in Appendix.

\subsection{Main Experiments}

\begin{figure*}[t]
	\vspace{-2em}
	\centering 	
	\subfloat[Contextual  Retrieval]	
	{\centering\includegraphics[width=0.33\linewidth]{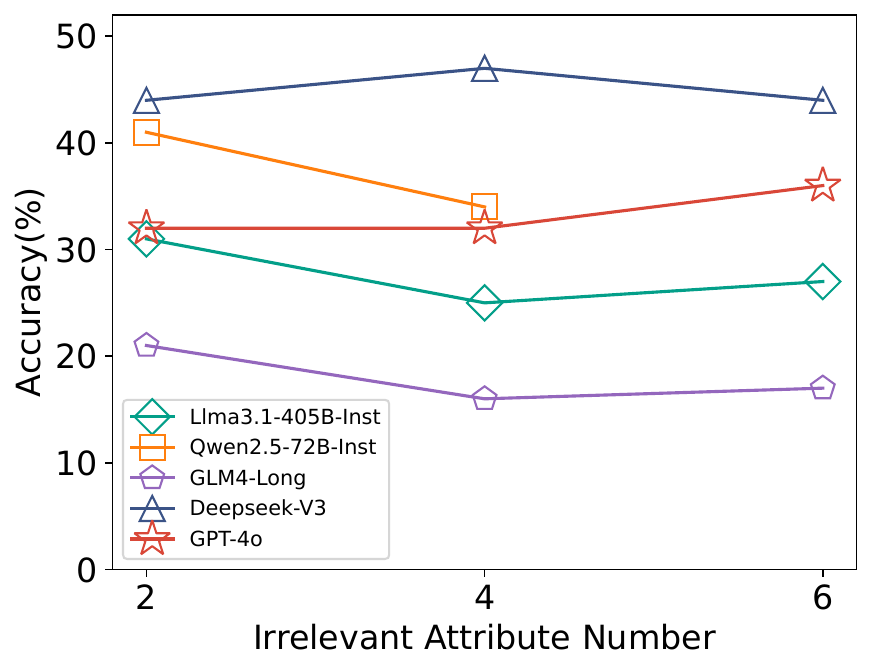}}
	\hfill
	\subfloat[Comparison]
	{\centering\includegraphics[width=0.33\linewidth]{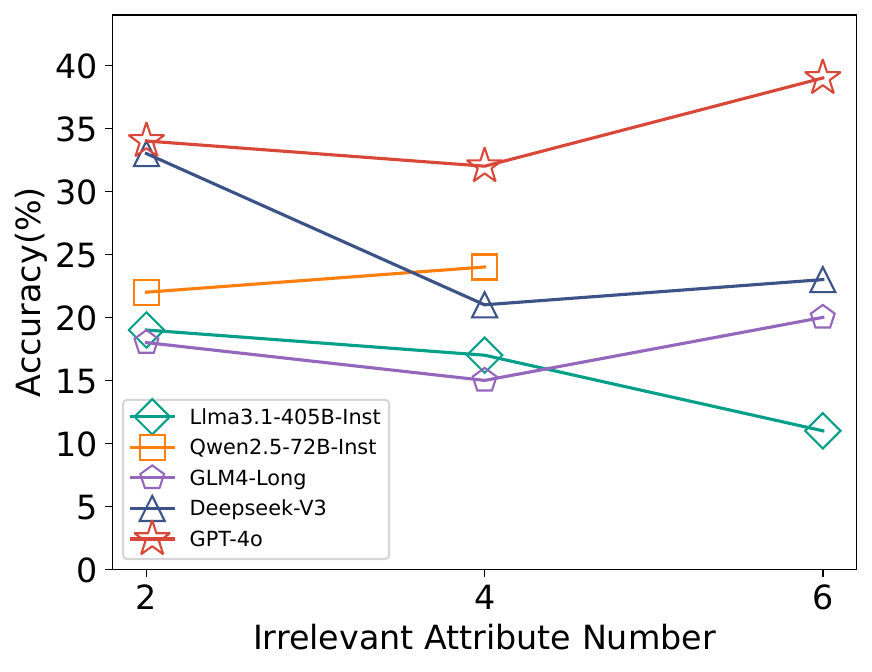}}
	\subfloat[Summary]	
	{\centering\includegraphics[width=0.33\linewidth]{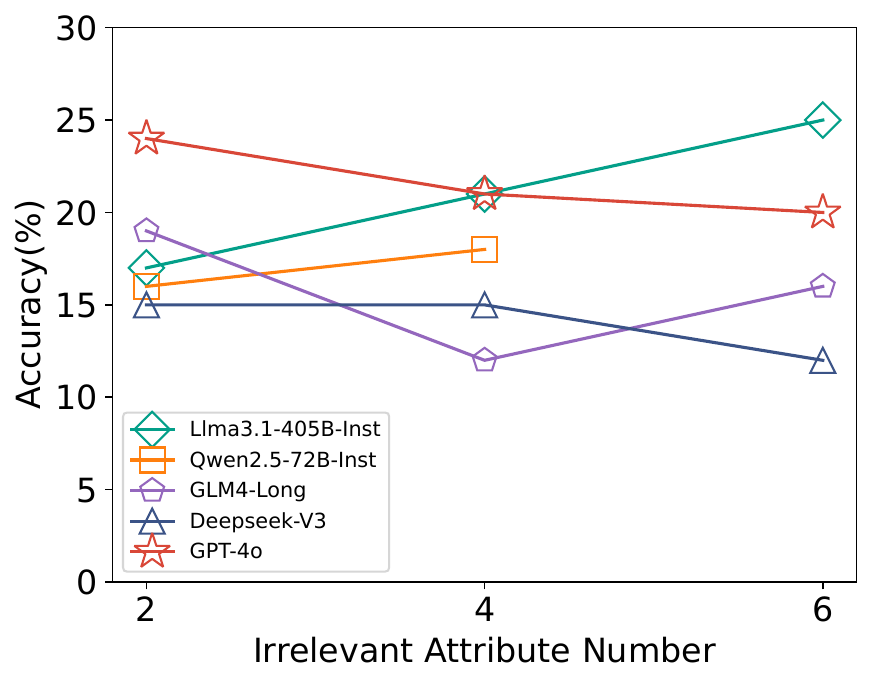}}
	\hfill
	
	\caption{Evaluation on noisy stock dataset. Due to the input sequence length limit of Qwen2.5-72B-Inst on the API platform, the data containing 6 irrelevant attributes cannot be evaluated using this model.}
	\label{fig:noisy_stock}
	
\end{figure*}

\begin{figure*}[t]
	\vspace{-1em}
	\centering 	
	\subfloat[Accuracy on $Q_{oper}$  (i.e., $a+b$)]	
	{\centering\includegraphics[width=0.32\linewidth]{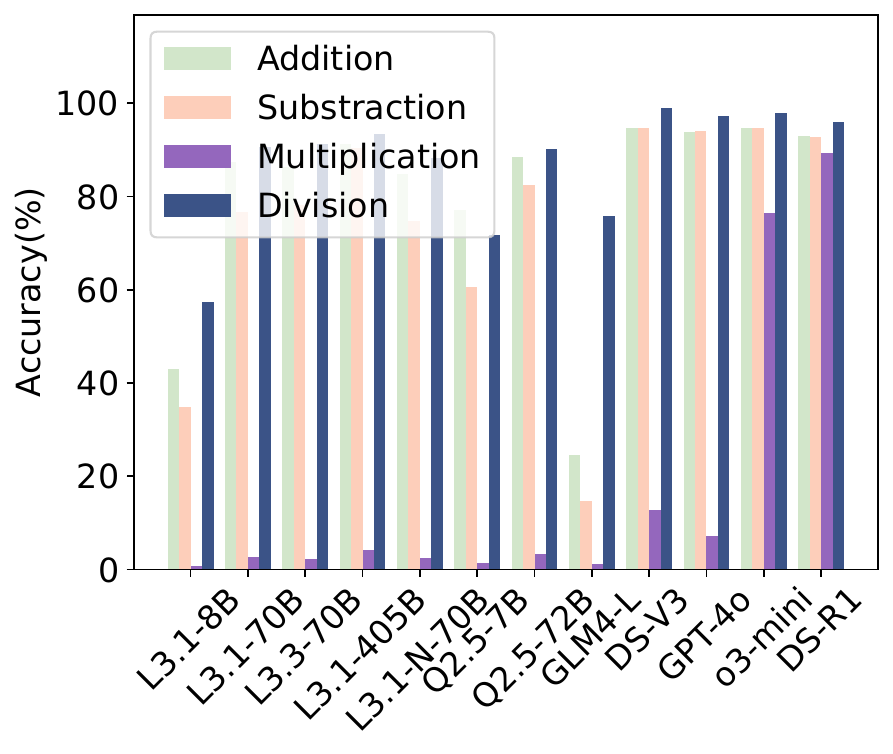}}
	\hfill
	\subfloat[ $Q_{oper}$  of different digits]
	{\centering\includegraphics[width=0.32\linewidth]{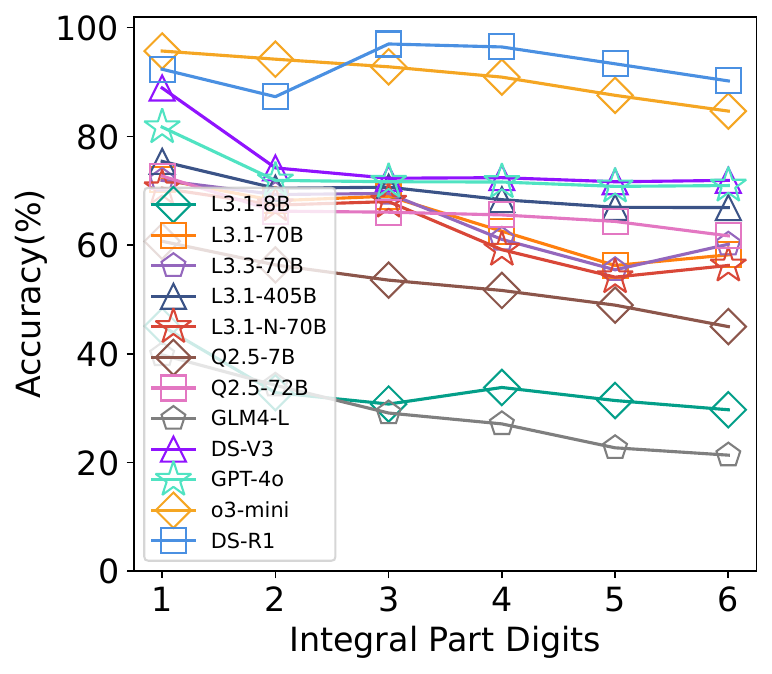}}
	\subfloat[Accuracy on $Q_{context}$  (i.e., $a$ plus $b$)]
	{\centering\includegraphics[width=0.32\linewidth]{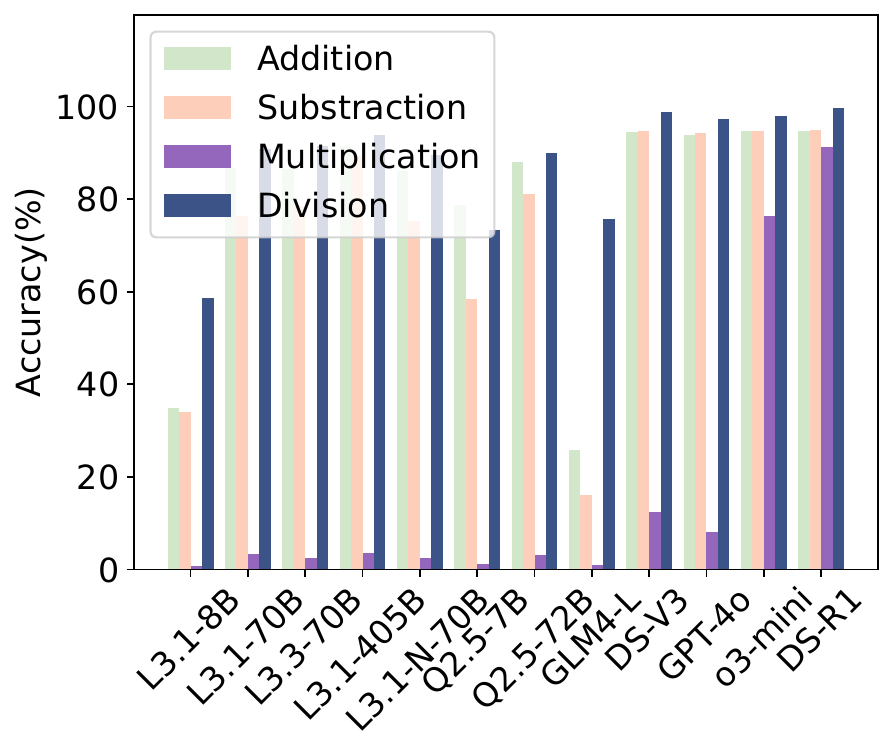}}
	
	\caption{Evaluation on arithmetic operation.}
	\label{fig:arithmetic_fig}
	\vspace{-1em}
\end{figure*}

\noindent \textbf{Evaluation on Contextual Retrieval, Comparison, Summary, and Logic Reasoning Abilities.}
As shown in Table~\ref{tab:main_experiments}, 
current popular and effective LLMs perform poorly on basic numerical tasks, 
including retrieval, comparison, summarization, and logic reasoning. 
The random baseline for each task is 12.5\%, as there are 8 choices, and the probability of randomly selecting the correct answer is 1/8. 
Human evaluation was conducted by three undergraduate students. 



Firstly, LLMs particularly struggle with accurately retrieving numerical data. This limitation arises from LLMs treating numbers as discrete tokens rather than continuous ones, coupled with insufficient exposure to structured numerical datasets during training, which restricts their ability to handle simple numeric retrieval tasks. 
Most LLMs rely on surface-level patterns, treating numbers as symbols without understanding their true magnitudes. Studies like MATH-Perturb~\cite{huang2025math} and GSM-Symbolic~\cite{mirzadeh2024gsm}  show that simple changes to numbers or variable names in math problems significantly reduce LLM accuracy, revealing their reliance on learned string patterns rather than genuine numerical understanding.
Secondly, LLMs demonstrate weaknesses in recognizing numerical relationships, such as greater-than or less-than comparisons, due to a lack of numerical semantics and underdeveloped arithmetic reasoning capabilities. 
Thirdly,
LLMs also perform poorly in summarizing numerical data (e.g., calculating sums or means), reflecting their inability to execute multi-step numerical operations. 
Similarly, logic reasoning tasks, especially those involving patterns or sequences, are particularly challenging, which are important for real-world applications.
These tasks require multi-step reasoning, pattern recognition, and arithmetic operations, which expose the architectural limitations of current LLMs.

\noindent \textbf{Evaluation on  Different Context Length.}
We evaluate LLMs on varying context lengths.
Specifically, we categorize the contexts of number lists, stock, and weather into short and long contexts.
The average token numbers for the short and long contexts across the three datasets are listed in Table~\ref{tab:data_stat_short_long}.
As illustrated in Figure~\ref{fig:length_number}, Figure~\ref{fig:length_stock}, and Figure~\ref{fig:length_weather},
LLMs generally achieve lower accuracy on long contexts compared to short contexts. This is because long contexts require the model to have a stronger ability to capture long-range dependencies.
Furthermore, if an LLM fails to perform well on short contexts, it is unlikely to achieve good results on long contexts. 
It highlights the importance of  LLMs in understanding numeric data.

\noindent \textbf{Evaluation on Noisy Context.}
\textcolor{black}{To evaluate the numerical abilities of LLMs in noisy context, we augmented the structured data with unrelated attributes while preserving the original relevant data. For instance, in the stock dataset, we added attributes such as Stock Price-to-Earnings, Dividend Per Share (DPS), and 52-Week High/Low to each instance. These attributes are unrelated to the user queries, and the amount of noise corresponds to the number of irrelevant attributes added. Specifically, we introduced $k \in \{2, 4, 6\}$ irrelevant attributes to each instance in both the stock and weather datasets. As shown in Figure~\ref{fig:noisy_stock} and Figure~\ref{fig:noisy_weather} in the Appendix, as $k$ increases, most LLMs exhibit degraded performance. This demonstrates that irrelevant context can negatively impact the numerical retrieval and reasoning abilities of LLMs.}


\noindent \textbf{Evaluation on Arithmetic Operations}
Similarly, 
we evaluate five strong LLMs on arithmetic operations.
Specifically, as illustrated in Figure~\ref{fig:arithmetic_fig}~(a), even for simple arithmetic operations involving two numbers, LLMs fail to achieve 100\% accuracy. 
Moreover, as the number of digits increases shown in Figure~\ref{fig:arithmetic_fig}~(b), the accuracy of LLMs decreases, highlighting their limited ability to handle arithmetic tasks effectively, which is also observed in~\citep{qiu2024dissecting}.
This poor performance stems from how LLMs generate responses. LLMs  predict the highest-order digit  before the lower-order digit~\citep{zhang2024reverse}, contradicting the standard arithmetic logic of progressing from lower- to higher-order digits.
In particular, Figure~\ref{fig:arithmetic_fig}~(a) and (c) shows that LLMs perform similarly on addition, subtraction, and division operations but achieve extremely low accuracy on multiplication tasks.

\begin{table}[]
	\centering
	\small
	\vspace{-1em}
	\caption{Evaluation on mixed-number-string data with lengths ranging from 50 (i.e., 50 L) to 200.}
	\vspace{-1em}
	\begin{tabular}{c|cccc}
		\toprule
		\textbf{Model}    & \textbf{50 L} & \textbf{100 L} & \textbf{150 L} & \textbf{200 L} \\ \midrule

		\textbf{\texttt{LLama3.1-405B }}& 10.8      & 9.2        & 3.2        & 2.2        \\  
		
		\textbf{\texttt{Qwen2.5-72B}}   & 3.0         & 1.2        & 0.6        & 0.2        \\  
		
		\textbf{\texttt{GLM4-Long}  }   & 6.6       & 4.8        & 3.0          & 2.4        \\  
		
		\textbf{\texttt{GPT-4o }}       & 18.2      & 6.4        & 4.0          & 4.2        \\ 
		
		\textbf{\texttt{DeepSeek-V3}}   & 13.2      & 4.0          & 3.2        & 2.0          \\  
		\midrule
		\textbf{\texttt{Human Eval } }   & \textbf{100}      & \textbf{100}        & \textbf{100}         & \textbf{100}        \\ \bottomrule
	\end{tabular}
\vspace{-1em}
	\label{tab:number_counting}
\end{table}

\noindent \textbf{Evaluation on Number Recognition via Mixed-number-string Dataset.}
We evaluate the number recognition ability of effective LLMs by identifying numbers from mixed-number-string sequences. For this evaluation, we select five  effective LLMs based on Table~\ref{tab:main_experiments}, including DeepSeek-v3, GLM-4-Long, Llama3.1-405B, and Qwen2.5-72B.
As shown in Table~\ref{tab:number_counting}, all LLMs achieve extremely low accuracy in counting numbers within strings. Moreover, as the length of the string increases from 50 to 100, the accuracy of the LLMs decreases further.
These results highlight that LLMs are significantly weak at distinguishing numbers from strings. The underlying reason is that current LLMs treat numbers as strings during training. 
This training paradigm inherently limits their ability to understand and process numbers effectively.
Also, the Tokenizer can split a single number into multiple tokens, which can negatively affect the numeric meaning of each number.

\begin{table}[h]
	\centering
	\footnotesize
	\setlength{\tabcolsep}{5pt} 
	\caption{\textcolor{black}{Chain of Thought Evaluations.}}
	\label{tab:cot}
	\begin{tabular}{c|ccc|ccc}
		\toprule
		\multirow{2}{*}{\textbf{Model}} & \multicolumn{3}{c|}{\textbf{Number List}} & \multicolumn{3}{c}{\textbf{Stock}} \\ \cmidrule{2-7}
		& R & C & S & R & C & S \\ \midrule
		\textbf{Base} & 44.4 & 31.5 & 26.3 & 19.4 & 13.8 & 16.8 \\\hline
		\textbf{Plain-CoT} & 65.2 & 39.4 & 29.4 & 24.8 & 27.7 & 16.6 \\
		\textbf{PS-CoT} & 65.4 & 40.0 & 26.4 & 24.3 & 16.7 & 15.8 \\
		\textbf{Table-CoT} & 65.8 & 38.4 & 29.0 & 27.6 & 29.1 & 9.6 \\ \bottomrule
	\end{tabular}
\end{table}
\subsection{LLMs with Advanced Techniques}
\noindent {\textbf{Chain of Thoughts.} We evaluate the impact of Chain of Thought (CoT) on improving the numeric reasoning abilities of LLMs. We use Llama-3.3-70B-Instruct as the backbone LLM for evaluations. Specifically, we incorporate representative CoT approaches, including Plain CoT~\cite{wei2023chainofthoughtpromptingelicitsreasoning}, Plan-and-Solve (PS)-CoT~\cite{wang2023planandsolvepromptingimprovingzeroshot}, and Table-CoT~\cite{jin2023tabcotzeroshottabularchain}, to enhance the reasoning capabilities of Llama-3.3-70B-Instruct.}
As shown in the Table~\ref{tab:cot},
these CoT techniques slightly improve LLM performance on NumericBench for simple tasks by refining outputs, but they significantly increase processing time due to longer outputs. 
However, CoT fails to enhance performance on complex stock summary tasks, where additional reasoning steps may introduce noise. This highlights the need to develop LLMs with stronger inherent numerical reasoning abilities.


\noindent \textbf{Few-shot Learning.}
We evaluate the impact of few-shot demonstrations on improving the numeric reasoning abilities of LLMs. Specifically, we use Llama-3.3-70B-Instruct as the backbone and employ one-shot and two-shot setups. As shown in Table~\ref{tab:few-shot-learning}, the base model performs best on simpler tasks, such as Number List Retrieval, Comparison, and Summary. Few-shot learning (one-shot and two-shot setups) shows potential for improving performance in more complex tasks, such as Stock Retrieval and  Summary. However, the performance gains from few-shot learning are inconsistent, suggesting that task complexity and the type of few-shot setup significantly influence the results.

\begin{table}[t]
	\centering
	\small
 \vspace{-1em}
	\setlength{\tabcolsep}{5pt} 
	\caption{Few-shot Learning Evaluations.}
	\vspace{-1em}
	\label{tab:few-shot-learning}
	\begin{tabular}{c|ccc|ccc}
		\toprule
		\multirow{2}{*}{\textbf{Model}} & \multicolumn{3}{c|}{\textbf{Number List}} & \multicolumn{3}{c}{\textbf{Stock}} \\ \cmidrule{2-7}
		& R & C & S & R & C & S \\ \midrule
		\textbf{Base} & 44.4 & 31.5 & 26.3 & 19.4 & 13.8 & 16.8 \\\hline
		\textbf{One-shot} & 43.3 & 26.0 & 21.3 & 23.3 & 13.8 & 18.5 \\
		\textbf{Two-shot} & 43.6 & 26.0 & 21.3 & 21.4 & 15.0 & 14.8 \\ \bottomrule
	\end{tabular}
\vspace{-1em}
\end{table}

\begin{table}[h]
	\centering
	\footnotesize
	
	\setlength{\tabcolsep}{2pt} 
	\caption{\textcolor{black}{Supervised Fine-tuning.}}
	\label{tab:qlora sft}
	\begin{tabular}{c|ccc|ccc|ccc}
		\toprule
		\multirow{2}{*}{\textbf{Model}} & \multicolumn{3}{c|}{\textbf{Number List}} & \multicolumn{3}{c|}{\textbf{Stock}} & \multicolumn{3}{c}{\textbf{Weather}} \\ \cmidrule{2-10}
		& R & C & S & R & C & S & R & C & S \\ \midrule
		
		\textbf{Base} & 24.4 & 21.1 & 29.5 & 14.0 & 9.3 & 16.8 & 17.0 & 10.8 & 11.1 \\
		\textbf{QLoRA} & 62.8 & 52.3 & 59.1 & 14.0 & 9.3 & 15.0 & 9.3 & 14.5 & 9.1 \\ \bottomrule
	\end{tabular}
\end{table}

\noindent \textbf{Supervised Fine-tuning.}
We use QLoRA \cite{dettmers2023qloraefficientfinetuningquantized} to fine-tune Llama-3.1-8B-Instruct on our number list, weather, and stock datasets with shorter lengths. Each dataset is divided into 70\% for training, 10\% for evaluation, and 20\% for testing. Table~\ref{tab:qlora sft} summarizes our results on the test sets of the split datasets.
As shown in Table~\ref{tab:qlora sft}, the fine-tuned model achieves improvements in retrieval, comparison, and summary tasks on the simple Number List dataset. However, for more complex real-world datasets, such as comparison and summary tasks on stock and weather data, the fine-tuning process does not yield improvements. This highlights the limitations of fine-tuning in enhancing LLM performance on complex, real-world data and underscores the necessity of developing LLMs with inherent numerical reasoning and understanding capabilities to handle such scenarios effectively.

\subsection{Discussions on Numeracy Gaps of LLMs}
In summary, extensive experimental results show that current state-of-the-art LLMs perform poorly on six fundamental numerical abilities.
Here we discuss five potential reasons behind their poor performance on numerical tasks.

\noindent \textbf{Tokenizer Limitation.}
LLMs use tokenizers to split input text into smaller units (tokens). Thus,
Numbers are split into chunks as strings, based on statistical patterns in the training data.
For example, $10000$ is split into $100$ and $00$ tokens\footnote{\url{https://gptforwork.com/tools/tokenizer}}.
These tokenizers do not consider the real meaning of numbers and continuous magnitude of numbers~\cite{wallace-etal-2019-nlp,singh2024tokenization}.
Thus, LLMs do not perform well on simple number retrieval and comparison tasks.
 Recent studies~\cite{sathe2024improvingselfconsistencyllms,shen2023positionaldescriptionmatterstransformers,yang2025numbercookbook} demonstrate that diverse tokenization methods can enhance LLMs' numerical understanding. 

\noindent \textbf{Training Corpora Limitation.}
LLMs are trained on extensive corpora, which also limits their ability to understand numerical-related symbols, such as $*$.
For example, the multiplication of 246 and 369 can be denoted as $246*369$.
However, $246*369$ may be interpreted as a password or encrypted text, since $*$ in text strings is often associated with encryption.
Consequently, enabling LLMs to accurately interpret arithmetic symbols and perform numerical reasoning remains an open problem, as their understanding of these symbols is heavily influenced by the statistical patterns and contexts encountered during training~\cite{razeghi2022impact}.

\noindent \textbf{Training Paradigm Limitation.}
The training of LLMs relies on the next-token prediction paradigm, which is inherently misaligned with the logic of numerical computation.
For example, when solving $16 + 56$ with the result being $72$, an LLM will first predict the highest-order digit of the answer (i.e., $7$) before predicting the lower-order digit (i.e., $2$). This approach contradicts the fundamental logic of arithmetic computation, which typically proceeds from the lower-order digit to the higher-order digit.
This discrepancy implies that LLMs effectively need to know the entire result upfront to generate digits sequentially in the correct order. As a result, LLMs struggle to perform well even on simple arithmetic operations.

\noindent \textbf{Positional Embedding Limitation.}
Note that LLMs incorporate positional embeddings for  tokens in sequence inputs. In arithmetic operations like $12 + 26$ and $26 + 12$, the order of the numbers does not affect the result. However, LLMs assign different positional embeddings to the number $12$ in each equation, as its position in the sequence differs. 
This lack of invariance in positional embeddings for numbers can influence the results.
Therefore, how to design the positional embedding that improves numerical ability of LLMs without affecting the text understanding  of LLMs is critical~\cite{mcleish2024transformers,golovneva2024contextual}.

\noindent \textbf{Transformer Architecture Limitation.}
LLMs use Transformer to process input sequence, which rely on pattern recognition rather than explicit algorithmic reasoning.
The computational power of transformers has upper bounds~\cite{merrill2023parallelism}. Considering the complexity of arithmetic operations in real-world applications, it still needs to be theoretically investigated whether transformers can perform well on numerical operations.

\section{Conclusion and Future Directions}
In this paper, we propose a comprehensive benchmark NumericBench   to evaluate the six  fundamental numerical  abilities of LLMs, including number recognition, arithmetic operations, contextual retrieval, comparison, summary, and logical reasoning. 
Our experiments reveal significant gaps in LLMs' numerical reasoning, as even state-of-the-art models like GPT-4o and DeepSeek-V3 struggle with simple arithmetic, number retrieval, and multi-step reasoning tasks. 
These shortcomings  arise from tokenization issues, training paradigms, and architectural limitations, underscoring the need for more numerically-aware modeling approaches.

To address these gaps, several future directions deserve exploration. First, developing numerically-aware tokenizers that treat numbers as continuous magnitudes can enable LLMs to better understand numerical concepts. Second, designing pretraining objectives specifically tailored to numerical reasoning, rather than relying solely on next-token prediction, can help models become more proficient at solving numerical problems. Third, incorporating structured numerical datasets during training can enhance real-world applicability by grounding models in accurate and practical numerical contexts. Finally, exploring suitable positional embeddings and hybrid symbolic-numeric architectures shows significant promise for improving the numerical capabilities of LLMs. 

%
%

\section*{Limitations}
There are two main limitations of this paper.
Firstly, the numerical tasks encountered in real-world scenarios are often far more complex and diverse compared to the six datasets proposed in NumericBench. 
Expanding the scope to include a broader range of numerical reasoning categories, such as traffic, would provide a more comprehensive assessment. 
Nevertheless,  our work can serve as a meaningful  point, highlighting  the current limitations of LLMs in numerical tasks.
We also analyze the potential reasons why LLMs struggle with numerical reasoning tasks, which can be attributed to the inherent limitations of transformer architectures and the next-token prediction objective.
We
hope it inspires further efforts to address these challenges and develop   more advanced LLMs with enhanced numerical  capabilities.

Secondly, although we evaluate twelve state-of-the-art LLMs, several newer LLMs and their variants, such as Claude and OpenAI o1 from major companies, are not included in our experiments. 
The reason for this exclusion is the expensive cost of accessing these model APIs. 
In brief, evaluating additional LLM variants across Claude, OpenAI, Mistral and GLM, typically requires a budget of nearly \textbf{\$15,000} US dollars.
Specifically, experiments on the datasets in Table~\ref{tab:main_experiments} require approximately 180 million tokens as inputs, while all left experiments (e.g., noisy contexts) require about 84 million tokens as inputs. 
For 1 million input tokens, Claude 3 Opus costs \$15\footnote{\url{https://www.anthropic.com/pricing\#anthropic-api}}, 
Claude 3.5 Sonnet costs \$3\footnote{\url{https://www.anthropic.com/pricing\#anthropic-api}}, 
OpenAI o1 costs \$15\footnote{\url{https://openai.com/api/pricing/}}, 
Gemini 1.5 Pro costs \$12.5\footnote{\url{https://ai.google.dev/pricing\#1_5pro}}, 
GLM4-Plus costs \$6.89\footnote{\url{https://bigmodel.cn/pricing}},
Mistral Large 24.11 costs \$2 \footnote{\url{https://mistral.ai/en/products/la-plateforme}},
and Mixtral 8x22B costs \$2 \footnote{\url{https://mistral.ai/en/products/la-plateforme}}.

If we conduct experiments above with these top-tier models from major companies, it would cost at least 3960 dollars for Claude 3 Opus, 3960 dollars for OpenAI o1, 3300 dollars for Gemini 1.5 Pro, 1819 dollars for GLM4-Plus, 792 dollars for Claude Sonnet 3.5, 528 dollars for Mistral Large 24.11 and 528 dollars for Mixtral 8x22B, which is beyond our expected total experiment cost.

Also, for models such as OpenAI o1, 
which require generating really longer outputs for reasoning purposes, the output length is often unpredictable, while the model charges for \$60 per million output tokens, making the experiments even more expensive and difficult to control.
Considering that GPT-4o, OpenAI o3-mini and DeepSeek-R1 represent the most state-of-the-art LLM models, we believe our evaluation can reflect the current numerical abilities of leading-edge LLMs.
Therefore, our evaluation highlights the weaknesses of LLMs in numerical abilities and serves as a bridge to inspire further research focused on improving the numerical capabilities of these models.

\section*{Ethics Statement}
This work does have  any ethical issues.

\section*{Acknowledegments}
Prof. Lei Chen’s work is  supported by National Key Research and Development Program of China Grant No. 2023YFF0725100, National Science Foundation of China (NSFC) under Grant No. U22B2060, Guangdong-Hong Kong Technology Innovation Joint Funding Scheme Project No. 2024A0505040012, the Hong Kong RGC GRF Project 16213620, RIF Project R6020-19, AOE Project AoE/E-603/18, Theme-based project TRS T41-603/20R, CRF Project C2004-21G, Guangdong Province Science and Technology Plan Project 2023A0505030011, Guangzhou municipality big data intelligence key lab, 2023A03J0012, Hong Kong ITC ITF grants MHX/078/21 and PRP/004/22FX, Zhujiang scholar program 2021JC02X170, Microsoft Research Asia Collaborative Research Grant, HKUST-Webank joint research lab and 2023 HKUST Shenzhen-Hong Kong Collaborative Innovation Institute Green Sustainability Special Fund, from Shui On Xintiandi and the InnoSpace GBA.
Prof. Qing Li is supported by the Hong Kong Research Grants Council under General Research Fund (project no. 15200023) and Research Impact Fund (project no. R1015-23).
Dr. Haoyang Li is supported by research funds P0052504 and P0053707.

\bibliographystyle{acl_natbib}
\bibliography{acl} 
\appendix
\clearpage
 
\definecolor{exampleblue}{RGB}{0, 114, 188} 
\definecolor{exampleborder}{RGB}{0, 114, 188} 
\definecolor{redtext}{RGB}{204, 0, 0}         

\section{Appendix}
\begin{figure*}[t]
	\centering	
	\vspace{-1em}
	\frame{
		\includegraphics[width = 0.9\textwidth]{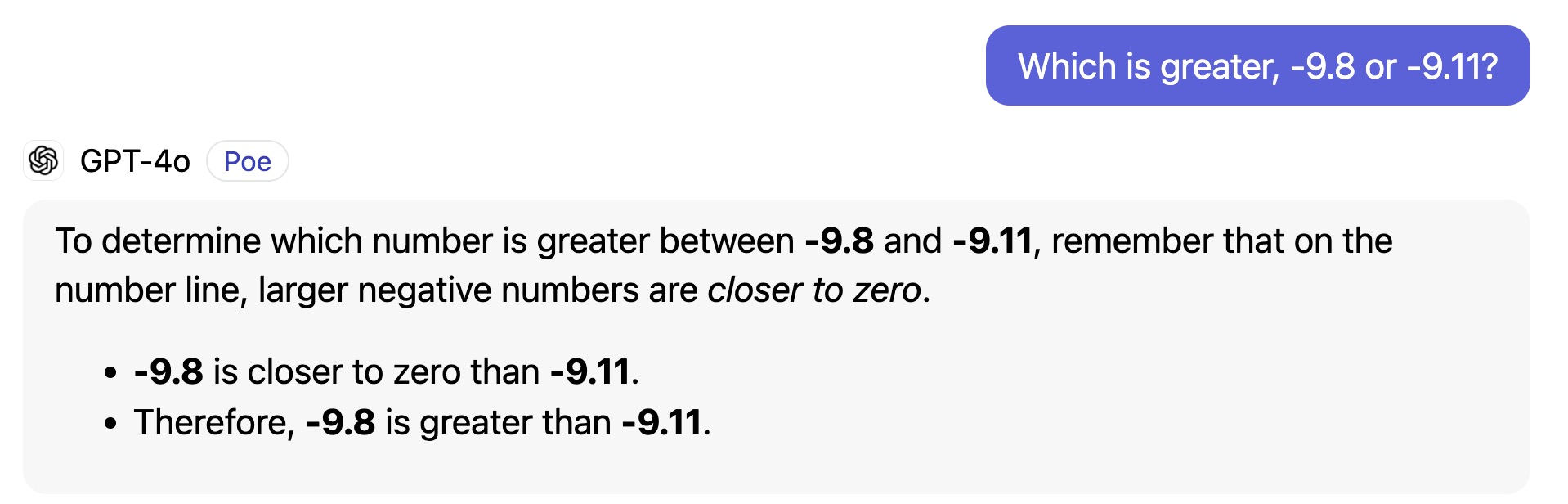}
	}
	\caption{Number comparisons on GPT-4o. The correct answer is -9.11. }
	\label{fig:number_compare}
\end{figure*}

\begin{figure*}[t]
	\centering	
	\vspace{-1em}
	\frame{
		\includegraphics[width = 0.9\textwidth]{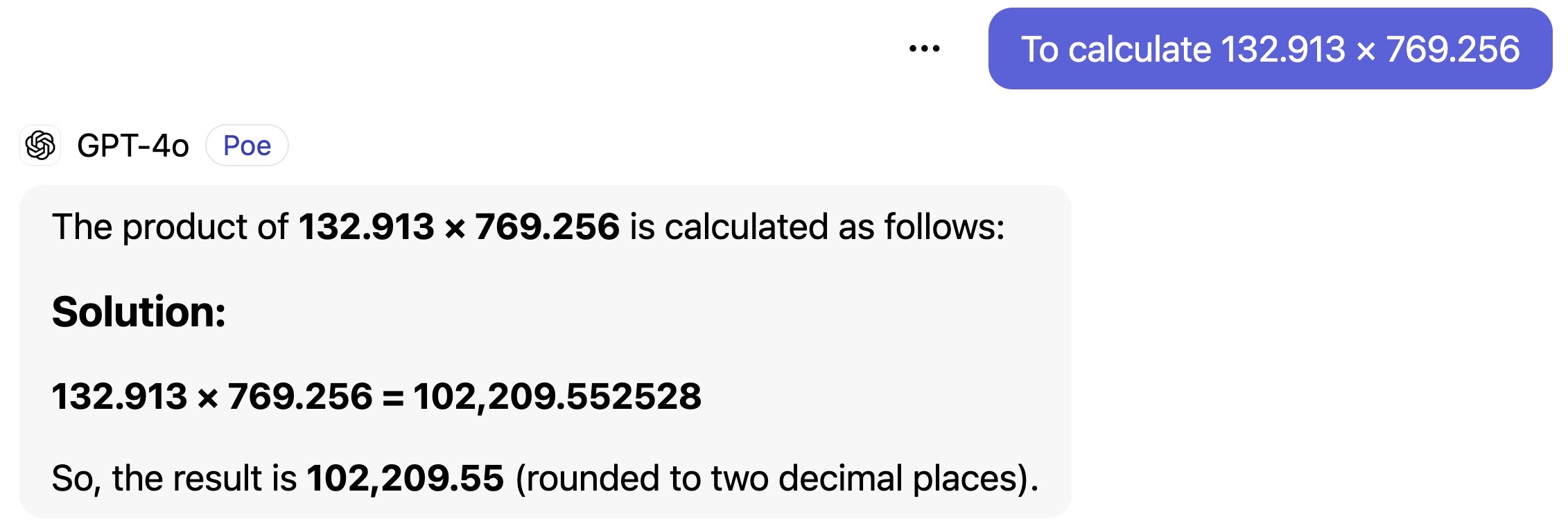}
	}
	\caption{Number multiplication on GPT-4o. The correct answer is 102244.12. }
	\label{fig:multiplication}
\end{figure*}

\begin{figure*}[t]
	\centering	
	\vspace{-1em}
	\frame{
		\includegraphics[width = 0.9\textwidth]{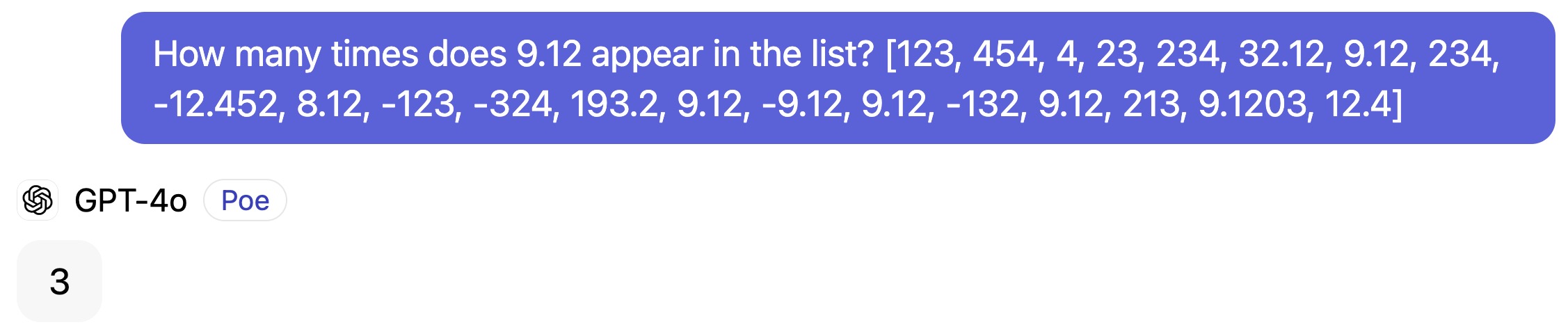}
	}
	\caption{Number counting on GPT-4o, which is required directly give answer. The correct answer is 4. }
	\label{fig:number_couting}
\end{figure*}

\begin{figure*}[t]
		\centering 	
		\subfloat[Contextual Retrieval]	
		{\centering\includegraphics[width=0.33\linewidth]{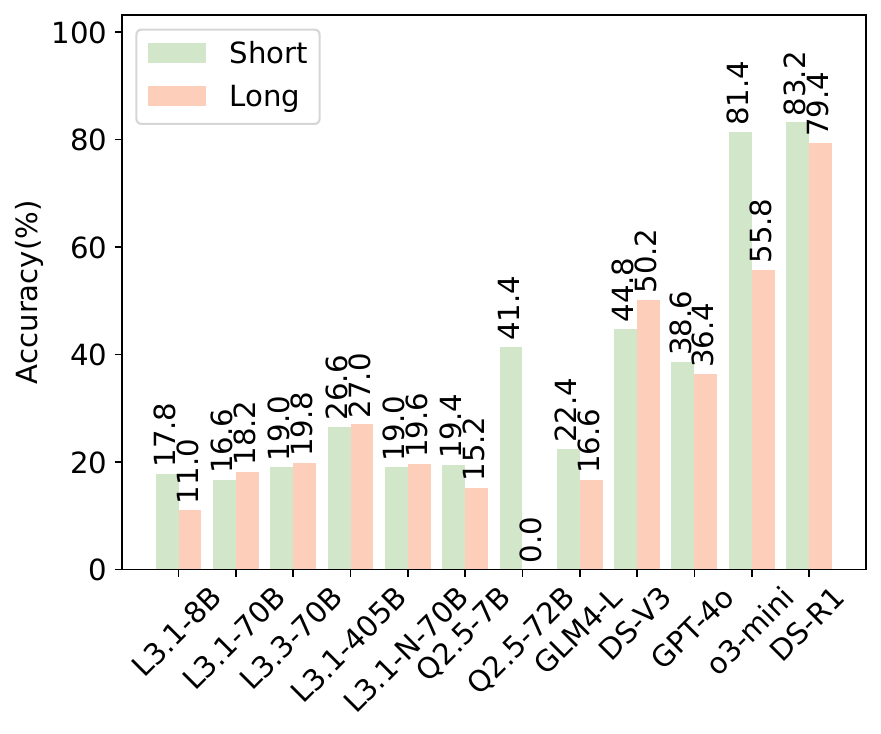}}
		\hfill
		\subfloat[Comparison]
		{\centering\includegraphics[width=0.33\linewidth]{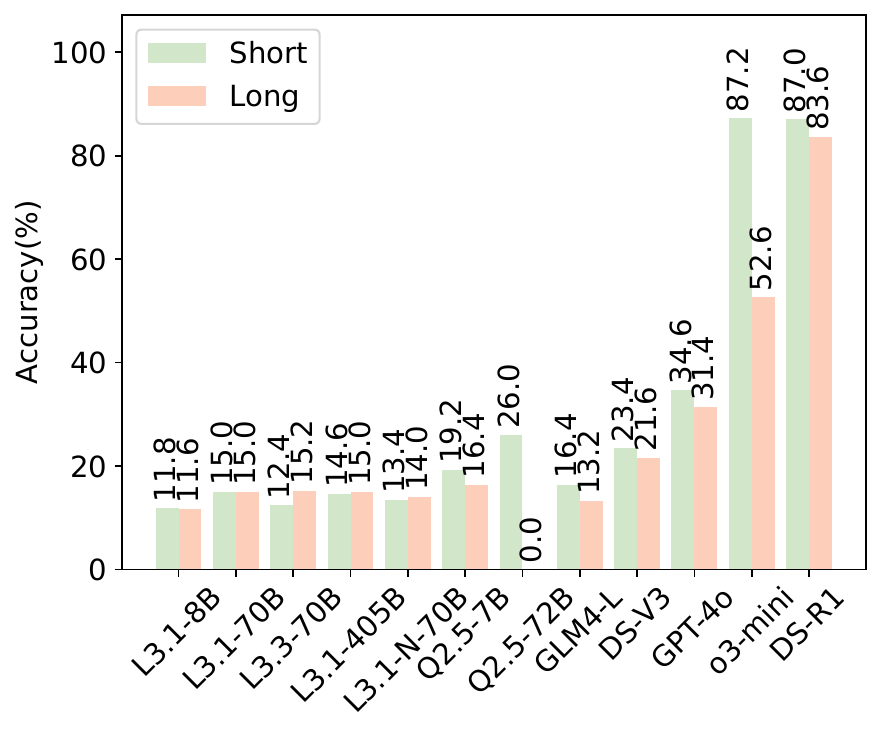}}
		\subfloat[Summary]	
		{\centering\includegraphics[width=0.33\linewidth]{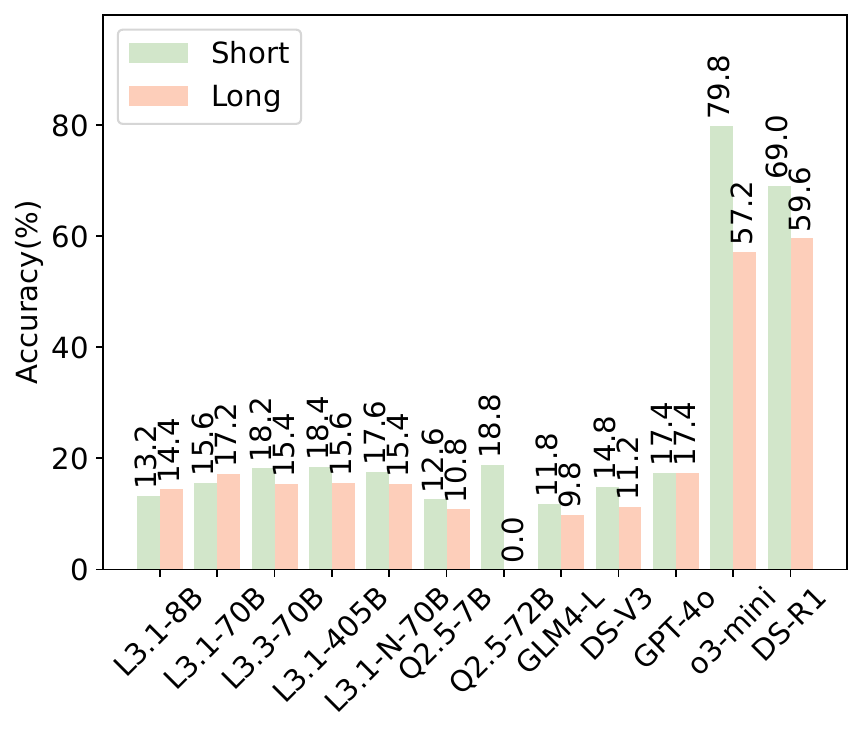}}
		\hfill
		%
		\caption{ {Evaluation on short and long context on stock dataset. Due to the input sequence length limit of Qwen2.5-72B-Inst on the API platform, the long dataset of all three abilities cannot be evaluated using this model.}}
		\label{fig:length_stock}
	\end{figure*}

\begin{figure*}[t]
		\centering 	
		\subfloat[Contextual Retrieval]	
		{\centering\includegraphics[width=0.33\linewidth]{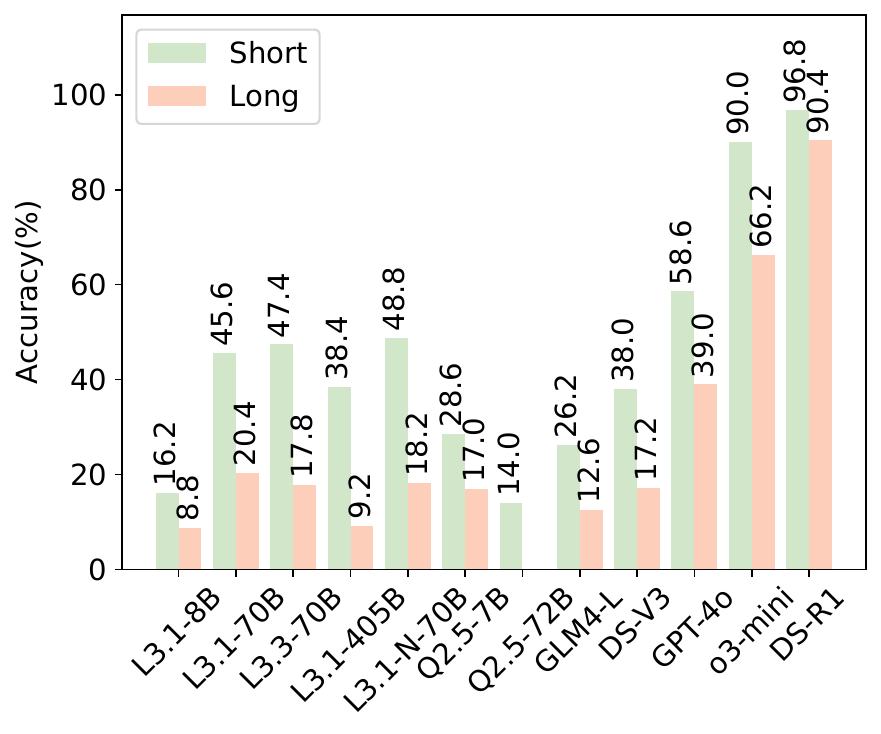}}
		\hfill
		\subfloat[Comparison]
		{\centering\includegraphics[width=0.33\linewidth]{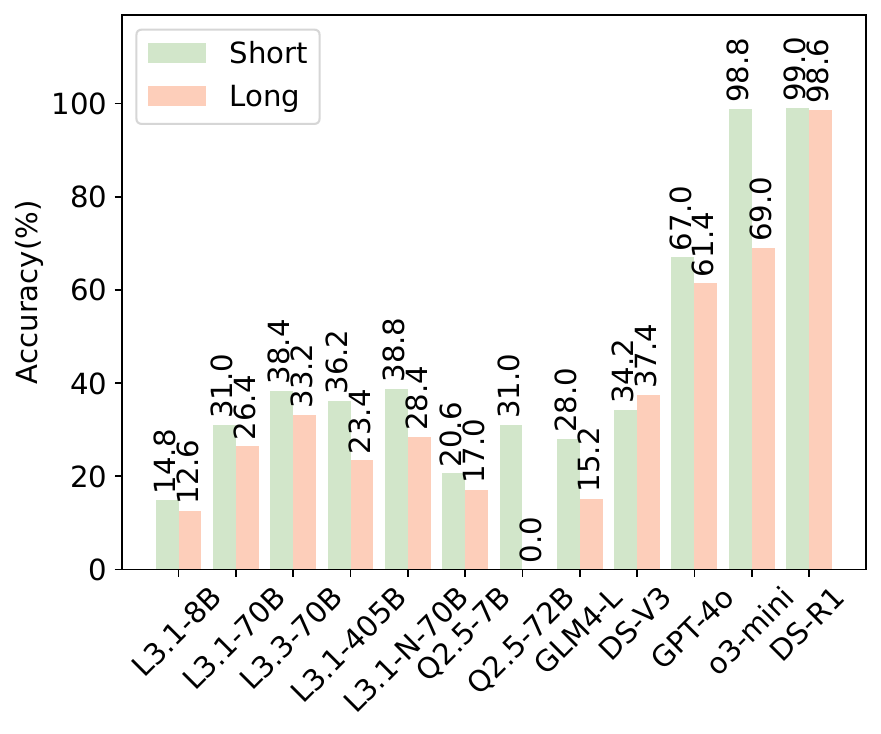}}
		\subfloat[Summary]	
		{\centering\includegraphics[width=0.33\linewidth]{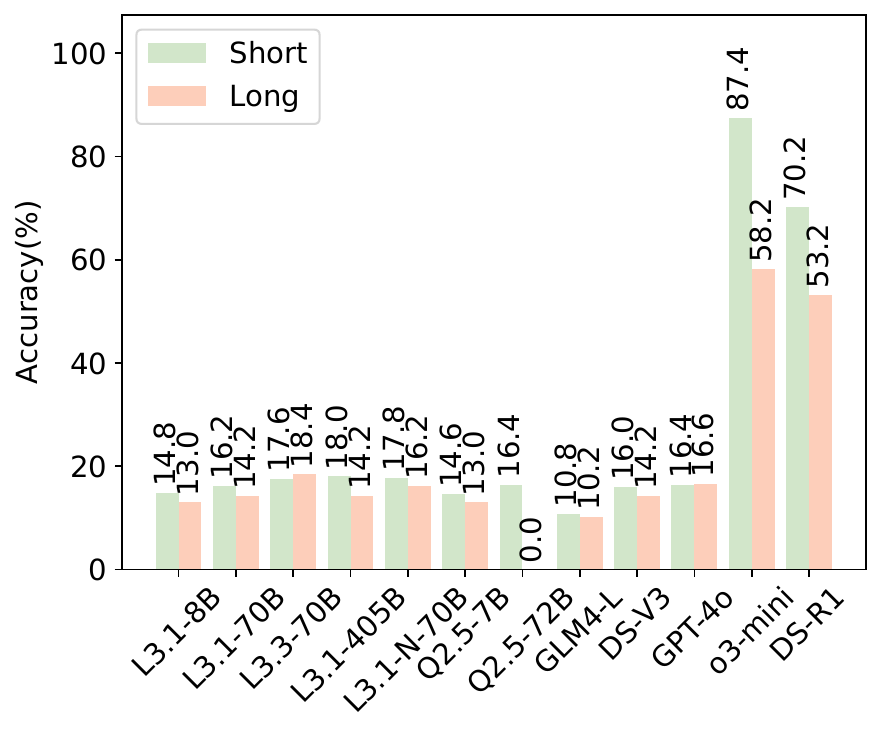}}
		\hfill
		%
		\caption{ {Evaluation on short and long context on weather dataset. Due to the input sequence length limit of Qwen2.5-72B-Inst on the API platform, the long dataset of all three abilities cannot be evaluated using this model.}}
		\label{fig:length_weather}
		
	\end{figure*}

\begin{figure*}[t]
		
		\centering 	
		\subfloat[Contextual  Retrieval]	
		{\centering\includegraphics[width=0.33\linewidth]{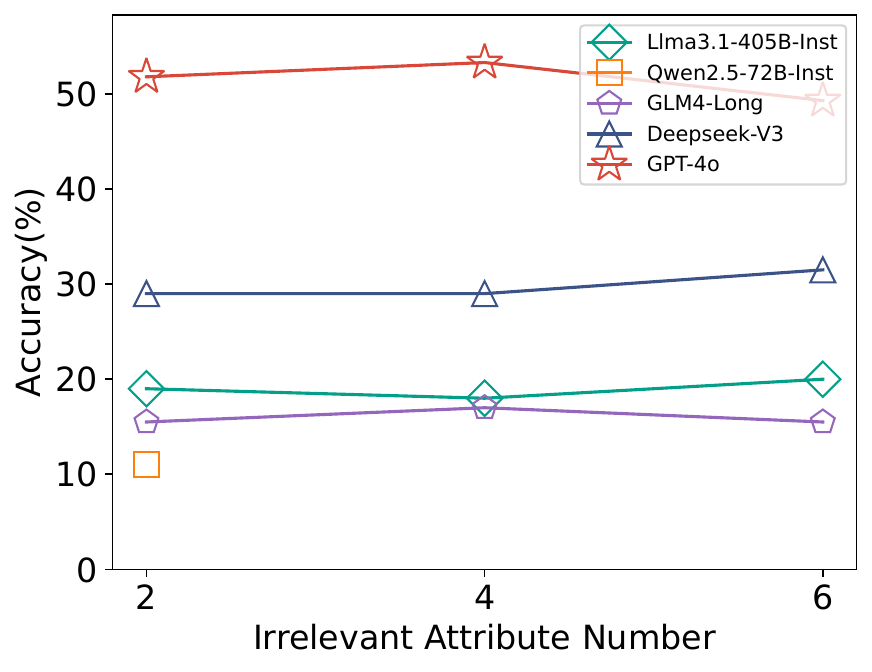}}
		\hfill
		\subfloat[Comparison]
		{\centering\includegraphics[width=0.33\linewidth]{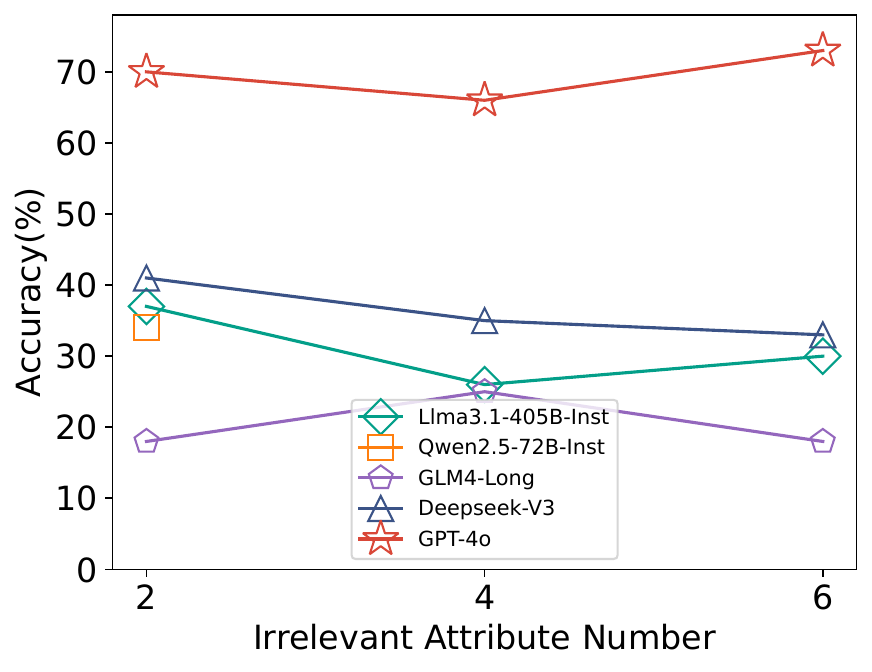}}
		\subfloat[Summary]	
		{\centering\includegraphics[width=0.33\linewidth]{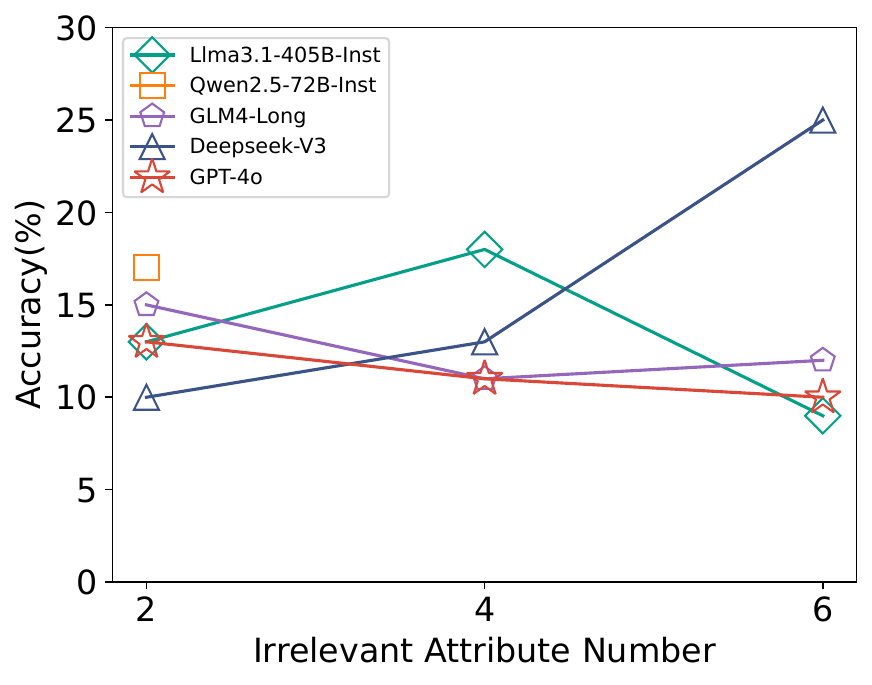}}
		\hfill
		%
		\caption{ {Evaluation on  noisy weather dataset. Due to the input sequence length limit of Qwen2.5-72B-Inst on the API platform, the data containing 4 and 6 irrelevant attributes cannot be evaluated using this model.}}
		\label{fig:noisy_weather}
		
	\end{figure*}

In this appendix, we provide additional details about the design of \textbf{NumericBench}, along with supplementary experimental results and case studies. The organization of the supplementary materials in this appendix is as follows:

\begin{enumerate}[leftmargin=*]
	\item {\textbf{Input format examples.}  
We transformed the structured JSON data into a string and appended it to the natural language question. The combined string, comprising the question and the structured data, was then used as the model's input. In Table~\ref{tab:input format}, we provide two examples.}
	
	\item \textbf{Question formats for contextual retrieval, comparison, and summary abilities.}  
As shown in Table~\ref{appx:number_question}, Table~\ref{appx:stock_question}, and Table~\ref{appx:weather_question}, 
we designed diverse question types tailored to each dataset to evaluate the three fundamental numerical abilities of LLMs: contextual retrieval, comparison, and summary. Contextual retrieval  assesses the model's capacity to accurately extract relevant numerical information from complex contexts; comparison tests the ability to analyze and compare numerical values; summary evaluates the synthesis of numerical information into concise and meaningful insights for tasks like reporting or trend analysis.

	\item \textbf{Basic numerical questions answered incorrectly by GPT-4o.}  
	As illustrated in Figure~\ref{fig:number_compare}, Figure~\ref{fig:multiplication}, and Figure~\ref{fig:number_couting}, GPT-4o failed to answer three basic numerical questions correctly. This result is surprising, considering GPT-4o's impressive performance in real-world applications. However, these findings highlight the weak fundamental numerical abilities of LLMs.
	
	\item \textbf{Token counts for short and long contexts.}  
	As shown in Table~\ref{tab:data_stat_short_long}, the token counts of long and short contexts differ significantly. 
	This distinction enables a more thorough evaluation of LLM performance across scenarios involving varying context lengths. Short contexts are designed to test the model's ability to process and understand concise information, focusing on immediate comprehension and reasoning. In contrast, long contexts present a more complex challenge, requiring the model to handle extended sequences of information, maintain coherence over a larger context window, 
	and retrieve relevant details from earlier parts of the input. Such two type length can more comprehensively evaluate LLMs. 
	
	\item \textbf{Additional experimental results on noisy and varying-length contexts.}  
	As shown in Figure~\ref{fig:length_stock} and Figure~\ref{fig:length_weather}, existing LLMs perform poorly on the stock and weather datasets, although they achieve better performance compared to their results on short contexts. 
	Similarly, as shown in Figure~\ref{fig:noisy_weather}, LLMs perform poorly on noisy weather data.
 
  \item \textbf{Output Token Statistics.} The number of tokens generated by LLMs is closely tied to their performance on NumericBench. Models augmented with CoT, as well as those inherently designed for reasoning, exhibit exceptional performance on our benchmark. However, this improvement comes with an increase in the number of output tokens. Table~\ref{tab:output token statistics} provides a detailed analysis of the token counts for the outputs of these high-performing models.
  
 \item \textbf{Real failure cases of math-oriented LLMs.} In this paper, we do not compare existing math-oriented LLMs, such as Metamath-Llemma-7B~\citep{yu2023metamath}, Deepseek-Math-7B-Instruct~\citep{deepseek-math}, InternLM2-Math-7B~\citep{ying2024internlmmathopenmathlarge}, and MAmmoTH-7B~\citep{yue2023mammoth}. 
 This is primarily because these math-oriented LLMs are designed for specialized geometric and structured mathematical problems. They are unable to understand the tasks in NumericBench, fail to follow a correct reasoning process, and directly produce meaningless outputs. These failure cases are illustrated in Figure~\ref{fig:fail_internlm}, Figure~\ref{fig:fail_ds_math}, Figure~\ref{fig:fail_llemma}, and Figure~\ref{fig:fail_mammoth}.

%
%
%
%
%
%
\end{enumerate}

\noindent \textbf{The Use of AI Tools.} When writing  this paper, we use Grammarly\footnote{https://www.grammarly.com/} for automated spell checking and use GPT-4o\footnote{https://platform.openai.com/docs/models/gpt-4o} to refine several sentences.

\clearpage

\begin{table*}[t]
	\caption{ {NumericBench Input Format. Due to the excessive length of the dataset input, the input examples provided in the table use \ldots to indicate the omission of certain parts.}}
	
	\label{tab:input format}
	\centering
	\renewcommand{\arraystretch}{1.15} 
	\setlength{\tabcolsep}{1.5pt} 
	\resizebox{\textwidth}{!}{
		\begin{tabular}{c|c}
		\toprule
		\textbf{Dataset} & \textbf{Input Example} \\ \midrule
		Number List &  {  \begin{tabular}[c]{@{}p{0.9\linewidth}@{}}
			You're an assistant designed to answer multiple choice questions, You'll be given some context and multiple choice questions about the context. For each question, you will only output the answer with the following format, without additional information such as how you solve the problem: The answer is (correct option). Option refers to capital letters like A, B, C, D, etc. Note that there is only one correct option. \\ \\
			The data you receive will contain a list of numbers. \\
			Question: Which index holds the greatest number in the list between the indices 3 and 14? Options: A: 3, B: 4, C: 6, D: 7, E: 9, F: 10, G: 12, H: 13 \\ \\
			Data: $[-778.124, -384366.23, -856, -243, -961 279, -10, \ldots]$
		\end{tabular}} \\ \midrule
		
		Stock &  {  \begin{tabular}[c]{@{}p{0.9\linewidth}@{}}
			You're an assistant designed to answer multiple choice questions, You'll be given some context and multiple choice questions about the context. For each question, you will only output the answer with the following format, without additional information such as how you solve the problem: The answer is (correct option). Option refers to capital letters like A, B, C, D, etc. Note that there is only one correct option.\\ \\
			This dataset contains key financial and trading information for stocks. The 'stock\_code' column represents the unique identifier for each stock, while 'date' indicates the trading date. 'close\_price' and 'open\_price' provide the close and open prices of the stock on the given trading day, respectively. 'quantity\_relative\_ratio' compares the day's trading volume to the average volume over a prior period \ldots \\
			Note that exceeding some number means greater than that number, while reaching some number means greater than or equal to that number.\\
			Question: How many days had a close price higher than the open price, with the quantity relative ratio exceeding 54\%?\\ \\
			Data: $[\{date: 2023.11.03, close\_price: 10.87, open\_price: 10.9, quantity\_rel_ratio: 0.63, \ldots\}, \ldots]$
		\end{tabular}} \\ \bottomrule
		\end{tabular}
	}
\end{table*}
\clearpage

\begin{table*}[]
	\centering
	
	\caption{Question format on number list dataset. R: contextual retrieval, C: comparison, S: summary. In the contextual retrieval task, a number $x$ is randomly selected from the given number list. For the comparison task, the $k$-th largest number is randomly generated within the range of one to the length of the number list. The indices $x$ corresponds to twenty percent of the length of the number list, while $y$ corresponds to eighty percent of the length. The number $z$ is randomly chosen within the range $(\min(\text{list}), \max(\text{list}))$. For the summary task, the top $k$ is set to thirty percent of the length of the number list.}
 
	\renewcommand{\arraystretch}{1.15}  
	\setlength{\tabcolsep}{1.5pt}  
	\begin{tabular}{c|c}
		\toprule
		\textbf{Ability}    & \textbf{Question Format} \\ \midrule
		\textit{\textbf{R}} &  \begin{tabular}[c]{@{}l@{}}
			$Q_0$: What is the index of the first occurrence of the number $x$ in the list?\\
			$Q_1$: What is the index of the last occurrence of the number $x$ in the list?\\
			$Q_2$: What is the number after the first occurrence of the number $x$ in the list?\\
			$Q_3$: What is the number before the last occurrence of the number $x$ in the list?\\
			$Q_4$: What is the index of the first even number in the list?\\
			$Q_5$: What is the index of the first odd number in the list?\\
			$Q_6$: What is the index of the last even number in the list?\\
			$Q_7$: What is the index of the last odd number in the list?
		\end{tabular} \\ \midrule
		\textit{\textbf{C}} &  \begin{tabular}[c]{@{}l@{}}
			$Q_8$: What is the index of the first occurrence of the $k$-th largest number in the given list?\\
			$Q_9$: Which index holds the greatest number in the list between the indices $x$ and $y$?\\
			$Q_{10}$: Which index holds the smallest number in the list between the indices $x$ and $y$?\\
			$Q_{11}$: Which number is closest to $z$ in the list between the indices $x$ and $y$?\\
			$Q_{12}$: Which number is furthest from $z$ in the list between the indices $x$ and $y$?\\
			$Q_{13}$: Which number is the largest among those less than $z$ in the list?\\
			$Q_{14}$: Which number is the smallest among those greater than $z$ in the list?
		\end{tabular} \\ \midrule
		\textit{\textbf{S}} &  \begin{tabular}[c]{@{}l@{}}
			$Q_{15}$: What is the maximum sum of any two consecutive items in the list?\\
			$Q_{16}$: What is the maximum sum of any three consecutive items in the list?\\
			$Q_{17}$: What is the maximum absolute difference between two consecutive items in the list?\\
			$Q_{18}$: What is the sum of the indices of the top $k$ largest numbers in the list?\\
			$Q_{19}$: What is the sum of the indices of the top $k$ smallest numbers in the list?\\
			$Q_{20}$: What is the average of the indices of the top $k$ largest numbers in the list?\\
			$Q_{21}$: What is the average of the indices of the top $k$ smallest numbers in the list?\\
			$Q_{22}$: How many times do numbers consecutively increase for more than five times?\\
			$Q_{23}$: How many times do numbers consecutively decrease for more than five times?\\
			$\cdots \cdots$ \\
		\end{tabular} \\ \bottomrule
	\end{tabular}	
	\label{appx:number_question}
\end{table*}
\clearpage

\begin{table*}[]
	
	\caption{Question format on stock dataset. R: contextual retrieval, C: comparison, S: summary. $x$ and $y$ lie within the minimum and maximum ranges of their respective attributes. The top $k$ corresponds to thirty percent of the number list. $date_1$ represents the day at the twentieth percentile of the stock history, while $date_2$ corresponds to the day at the eightieth percentile.}
	\centering
	\renewcommand{\arraystretch}{1.15} 
	\setlength{\tabcolsep}{1.5pt} 
	\begin{tabular}{c|c}
		\toprule
		\textbf{Ability}    & \textbf{Question Format} \\ \midrule
		\textit{\textbf{R}} &  \begin{tabular}[c]{@{}l@{}}
			$Q_0$: On which date did the close price of the stock first reach $x$ yuan?\\
			$Q_1$: On which date did the highest price of the stock first reach $x$ yuan?\\
			$Q_2$: On which date did the volume of the stock first reach $x$ lots?\\
			$Q_3$: On which date did the value of the stock first reach $x$ thousand yuan?\\
			$Q_4$: On which date did the price change rate of the stock first reach $x$\%?\\
			$Q_5$: On which date did the price change of the stock first reach $x$ yuan?\\
		\end{tabular} \\ \midrule
		\textit{\textbf{C}} &  \begin{tabular}[c]{@{}l@{}}
			\begin{tabular}[c]{@{}l@{}}
				$Q_6$: On which date did the stock have the highest turnover rate when the close \\price was greater than $x$ yuan?
			\end{tabular}\\
			
			\begin{tabular}[c]{@{}l@{}}
				$Q_7$: On which date did the stock have the highest quantity relative ratio when \\the open price was less than $x$ yuan?
			\end{tabular}\\
			
			\begin{tabular}[c]{@{}l@{}}
				$Q_8$: On which date did the stock have the highest difference between the highest \\and lowest prices when the trading volume exceeded $x$ lots?
			\end{tabular}\\
			
			\begin{tabular}[c]{@{}l@{}}
				$Q_9$: On which date did the stock record the highest daily average price, calculated \\as 'value' divided by 'volume,' when the PE ratio was less than $x$?
			\end{tabular}\\
			
			\begin{tabular}[c]{@{}l@{}}
				$Q_{10}$: Among the top-$k$ trading value days, on which date did the stock have the \\lowest close price?
			\end{tabular}\\
			
			\begin{tabular}[c]{@{}l@{}}
				$Q_{11}$: When the quantity relative ratio exceeded $x$, on which date did the stock \\have the highest sum of the open price and close price?
			\end{tabular}\\
			
			\begin{tabular}[c]{@{}l@{}}
				$Q_{12}$: When the absolute price change rate exceeded $x$\%, on which date did the \\stock have the highest difference between the highest and lowest prices?
			\end{tabular}
		\end{tabular} \\ \midrule
		\textit{\textbf{S}} &  \begin{tabular}[c]{@{}l@{}}
			\begin{tabular}[c]{@{}l@{}}
				$Q_{13}$: How many days had a volume greater than $x$ from $date_1$ to $date_2$?
			\end{tabular}\\
			
			\begin{tabular}[c]{@{}l@{}}
				$Q_{14}$: How many days had the close price higher than the open price from \\$date_1$ to $date_2$?
			\end{tabular}\\
			
			\begin{tabular}[c]{@{}l@{}}
				$Q_{15}$: How many days had a close price higher than the open price, with the \\quantity relative ratio exceeding $x$\%?
			\end{tabular}\\
			
			\begin{tabular}[c]{@{}l@{}}
				$Q_{16}$: How many days had the close price reach $x$ yuan with the absolute price \\change rate exceeding $x$\%?
			\end{tabular}\\
			
			\begin{tabular}[c]{@{}l@{}}
				$Q_{17}$: What was the average trading volume when both the turnover rate \\exceeded $x$\% and the price change rate was greater than $y$\%?
			\end{tabular}\\
			
			\begin{tabular}[c]{@{}l@{}}
				$Q_{18}$: Excluding non-trading days, how many times did the open price of \\the stock rise for three or more consecutive days?
			\end{tabular}\\
			
			\begin{tabular}[c]{@{}l@{}}
				$Q_{19}$: Excluding non-trading days, how many times did the close price of \\the stock rise for three or more consecutive days?
			\end{tabular}\\
			
			\begin{tabular}[c]{@{}l@{}}
				$Q_{20}$: Excluding non-trading days, how many times did the open price and \\close price of the stock both rise for three or more consecutive days?
			\end{tabular}\\
		
			\begin{tabular}[c]{@{}l@{}}
			$\cdots \cdots$
		\end{tabular}
		
		\end{tabular} \\ \bottomrule
	\end{tabular}
\label{appx:stock_question}
\end{table*}
\clearpage

\begin{table*}[]
	\centering
	\caption{Question format on weather dataset.  R: contextual retrieval, C: comparison, S: summary. The value of $x$ falls within the minimum and maximum ranges of its respective attribute. $date_1$ represents the day at the twentieth percentile of the stock history, while $date_2$ represents the day at the eightieth percentile.}
	\begin{tabular}{c|c}
		\toprule
		\textbf{Ability}    & \textbf{Question Format} \\ \midrule
		\textit{\textbf{R}} &  \begin{tabular}[c]{@{}l@{}}
			 {$Q_0$: On which date did the temperature at two meters lastly reach $x$°C?}\\
			 {$Q_1$: On which date did the relative humidity at two meters lastly exceed $x$\%?}\\
			 {$Q_2$: On which date did the dew point temperature at two meters lastly drop below $x$°C?}\\
			 {$Q_3$: On which date did the precipitation lastly exceed $x$ mm?}\\
			 {$Q_4$: On which date did the sea-level air pressure lastly exceed $x$ hPa?}\\
			 {$Q_5$: On which date did the cloud cover lastly reach $x$\%?}\\
			 {$Q_6$: On which date did the wind speed at 10 meters lastly exceed $x$ m/s?}
		\end{tabular} \\ \midrule
		\textit{\textbf{C}} &  \begin{tabular}[c]{@{}l@{}}
			\begin{tabular}[c]{@{}l@{}}
				$Q_7$: On which date did the temperature at two meters reach its highest value \\ 
				when the relative humidity was below $x$\%? 
			\end{tabular} \\
			
			\begin{tabular}[c]{@{}l@{}}
				$Q_8$: On which date did the relative humidity at two meters reach its lowest value \\ 
				when the temperature at two meters was above $x^\circ$C?
			\end{tabular} \\
			
			\begin{tabular}[c]{@{}l@{}}
				$Q_9$: On which date did the difference between the temperature and dew point \\ 
				at two meters reach its maximum when the cloud cover was below $x$\%? 
			\end{tabular} \\
			
			\begin{tabular}[c]{@{}l@{}}
				$Q_{10}$: On which date did the precipitation reach its highest value \\ 
				when the temperature at two meters was below $x^\circ$C? 
			\end{tabular} \\
			
			\begin{tabular}[c]{@{}l@{}}
				$Q_{11}$: On which date did the cloud cover reach its lowest value \\ 
				when the wind speed at 10 meters exceeded $x$ m/s? 
			\end{tabular} \\
			
			\begin{tabular}[c]{@{}l@{}}
				$Q_{12}$: On which date did the wind speed at 10 meters reach its highest value \\ 
				when the precipitation exceeded $x$ mm? 
			\end{tabular} \\
			
			\begin{tabular}[c]{@{}l@{}}
				$Q_{13}$: On which date did the sea-level air pressure reach its highest value \\ 
				when the cloud cover was below $x$\%? 
			\end{tabular}
		\end{tabular} \\ \midrule
		\textit{\textbf{S}} &  \begin{tabular}[c]{@{}l@{}}
			\begin{tabular}[c]{@{}l@{}}
				$Q_{14}$: How many days had a temperature at two meters greater than $x^\circ$C \\from $date_1$ to $date_2$? 
			\end{tabular} \\
			
			\begin{tabular}[c]{@{}l@{}}
				$Q_{15}$: How many days had a relative humidity at two meters exceeding $x$\% \\from $date_1$ to $date_2$? 
			\end{tabular} \\
			
			\begin{tabular}[c]{@{}l@{}}
				$Q_{16}$: How many days had a precipitation greater than $x$ mm from $date_1$ \\to $date_2$? 
			\end{tabular} \\
			
			\begin{tabular}[c]{@{}l@{}}
				$Q_{17}$: What was the average temperature at two meters when the relative \\humidity exceeded $x$\%? 
			\end{tabular} \\
			
			\begin{tabular}[c]{@{}l@{}}
				$Q_{18}$: What was the average wind speed at 10 meters when the precipitation \\exceeded $x$ mm? 
			\end{tabular} \\
			
			\begin{tabular}[c]{@{}l@{}}
				$Q_{19}$: How many times did the temperature at two meters rise for three or more \\consecutive days? 
			\end{tabular} \\
			
			\begin{tabular}[c]{@{}l@{}}
				$Q_{20}$: How many times did the relative humidity at two meters drop for \\three or more consecutive days? 
			\end{tabular} \\

		\begin{tabular}[c]{@{}l@{}}
			$\cdots \cdots$
		\end{tabular} \\
	
		\end{tabular} \\ \bottomrule
	\end{tabular}
\label{appx:weather_question}
\end{table*}

\clearpage


\clearpage

\clearpage

%
%
	
	\begin{table*}[]
		\caption{The average token number on short and long instances for each data.}
		\centering
		\renewcommand{\arraystretch}{1.5} 
		\setlength{\tabcolsep}{3pt} 
		\begin{tabular}{c|c|cc|cc}
			\toprule
			\multirow{2}{*}{\textbf{Dataset}}                                               & \multirow{2}{*}{\textbf{Ability}} & \multicolumn{2}{c|}{\textbf{Short}}                            & \multicolumn{2}{c}{\textbf{Long}}                             \\ \cmidrule{3-6} 
			
			&                                   & \multicolumn{1}{c|}{\textbf{\# Instance}} & \textbf{Avg Token} & \multicolumn{1}{c|}{\textbf{\# Instance}} & \textbf{Avg Token} \\ \midrule
			
			\multirow{3}{*}{\textbf{\begin{tabular}[c]{@{}c@{}}Number\\ List\end{tabular}}} & \textit{Contextual Retrieval}

                  & \multicolumn{1}{c|}{500}                     &        809.12     & \multicolumn{1}{c|}{500}                     &         6599.34      \\   
			
			& \textit{Comparison}                        & \multicolumn{1}{c|}{500}                     &     804.86     & \multicolumn{1}{c|}{500}                     &        6566.27      \\ 
			
			& \textit{Summary}

                  & \multicolumn{1}{c|}{500}                     &       822.49      & \multicolumn{1}{c|}{500}                     &       6487.07       \\ \midrule
			
			\multirow{3}{*}{\textbf{Stock}}                                                 & \textit{Contextual Retrieval}

                  & \multicolumn{1}{c|}{500}                     &        18529.07      & \multicolumn{1}{c|}{500}                     &      36641.63     \\  
			& \textit{Comparison}                        & \multicolumn{1}{c|}{500}                     &    18539.58     & \multicolumn{1}{c|}{500}                     &      36651.22      \\ 
			& \textit{Summary}

                  & \multicolumn{1}{c|}{500}                     &      18504.51      & \multicolumn{1}{c|}{500}                     &       36618.07      \\ \midrule
			
			\multirow{3}{*}{\textbf{Weather}}                                               & \textit{Contextual Retrieval}

                  & \multicolumn{1}{c|}{500}                     &        18362.38        & \multicolumn{1}{c|}{500}                     &        36356.13    \\  
			& \textit{Comparison}                        & \multicolumn{1}{c|}{500}                     &        18371.11    & \multicolumn{1}{c|}{500}                     &       36365.27     \\ 
			& \textit{Summary}

                  & \multicolumn{1}{c|}{500}                     &        18334.15     & \multicolumn{1}{c|}{500}                     &        36328.27    \\ \bottomrule
		\end{tabular}
		\label{tab:data_stat_short_long}
	\end{table*}

	\begin{table*}[t]
		\centering
		\footnotesize
		\renewcommand{\arraystretch}{1.5} 
		\setlength{\tabcolsep}{3pt} 
		\caption{ {Output Token Statistics. We utilize Llama-3.3-70B-Instruct as our base model.}}
		\label{tab:output token statistics}
		\begin{tabular}{c|cc|cc|cc|c}
			\toprule
			\multirow{2}{*}{\textbf{Model}} & \multicolumn{2}{c|}{\textbf{Number List}} & \multicolumn{2}{c|}{\textbf{Stock}} & \multicolumn{2}{c|}{\textbf{Weather}}& \multirow{2}{*}{\textbf{Sequence}} \\ \cmidrule{2-7}
			& Short & Long & Short & Long & Short & Long & \\ \midrule
			Base & 5.00 & 5.00 & 5.00 & 5.00 & 5.00 & 5.09 & 5.00 \\
			Plain-CoT & 762.65 & 1386.60 & 456.45 & 402.18 & - & - & 1048.14 \\
			PS-CoT & 928.01 & 2776.77 & 512.37 & 515.25 & - & - & 913.09 \\
			Table-CoT & 872.65 & 1426.27 & 495.29 & 478.47 & - & - & 981.95 \\
			o3-mini & 2966.66 & 4158.25 & 3898.36 & 3907.98 & 2655.78 & 3466.44 & 4400.85 \\
			Deepseek-R1 & 5427.32 & 10463.43 & 4559.96 & 4837.71 & 2228.91 & 2482.29 & 8057.13 \\ \bottomrule
		\end{tabular}
	\end{table*}
	\clearpage
%
%

	\newtcolorbox[auto counter, number within=section]{examplebox}[2][]{%
		colback=white,        
		colframe=exampleborder, 
		coltitle=white,       
		fonttitle=\bfseries,  
		colbacktitle=exampleblue, 
		title=Example~\thetcbcounter: #2, 
	}
	\begin{figure*}
		\begin{examplebox}{InternLM2-Math-7B}
			\textbf{Failure:} \\Model repeatedly output the same number until the output sequence ends.
			
			\vspace{0.2cm}
			\textbf{User:} \\ Question: What is the maximum sum of any three consecutive items in the list? Options: A: 21501.9737231253, B: 21502.8144848529, C: 21501.9642201794, D: 21502.7354388437, E: 21502.0327435643, F: 21501.7526575621, G: 21501.7057960155, H: 21502 Data: [-5279, ....(100 numbers)]
			
			\vspace{0.3cm}
			\textbf{InternLM2-Math-7B Model:} \\
			Step 1: To find the maximum sum of any three consecutive items in the list, we need to identify the three largest numbers in the list and add them together. <PRM +> \\Step 2: We can start by sorting the list in ascending order to make it easier to identify the largest numbers. <PRM +> \\Step 3: After sorting, the list becomes: [-8509, -8509, -8509, -8509, -8509, -8509, -8509, -8509, -8509, -8509, -8509, -8509, -8509, -8509, -8509, -8509, -8509, -8509, -8509, -8509, -8509, -8509, -8509, -8509, -8509, -8509, -8509, -8509...(output too long)
		\end{examplebox}
		\caption{Failure type and conversation record of InternLM2-Math-7B. The model constantly generate the same "-8509" text until the output sequence was too long and truncated, leaving the problem unsolved.}
		\label{fig:fail_internlm}
	\end{figure*}
	
	\begin{figure*}
		\begin{examplebox}{DeepSeek-Math-7B}
				\textbf{Failure:} \\ Model outputs sequence without meaning, probably due to overly long input sequences. 
				
				\vspace{0.2cm}
				\textbf{User:} \\Question: What is the sum of the index of top 300 smallest numbers in the list? Options: A: 146734, B: 146738, C: 146733, D: 146724, E: 146728, F: 146727, G: 146721, H: 146729 Data: [3640, 1560, ...(100 numbers)]
				
				\vspace{0.3cm}
				\textbf{Deepseek-Math-7B Model:} \\
				31,3,9,3,1,1,3,3,3,3,3,3,3,3, 3,3,1,3,3,3,1,3,3,3, 3, 3,3,3, 1, 1,3,3, 3,1,3, 3,3,3, 1, 3,3,3, 3,3,1,3,1,3,91, 3,3,3, 3,31,3,3,3, 3, 339, 3, 3, 311, 1, 3, 1,3,3, 3,3,6,3,3, 3,31,3,3,1, 1, 1, 1, 3,1,1, 3,3, 1,1, 3,1,3,1, 1, 1, 1, 1, 19,1, 1, 1,191,1, 1,1, 19, 31,1,1, 1, 3,1, 1,3, 1133,3, 1, 1,1, 11, 11,3, 31, 3, 3,1,1,3,1, 1,33, 11,3,1911,3, 1,13,3131311119191,1, 3,1,11,1,19,11,1,1,1....(keeps outputting numbers)
		\end{examplebox}
		\caption{Failure type and conversation record of DeepSeek-Math-7B. The model only generates a series of numbers throughout its output, and we're unable to acquire any useful information from the output.}
		\label{fig:fail_ds_math}
	\end{figure*}	

	\begin{figure*}
		\begin{examplebox}{Metamath-Llemma-7B}
			\textbf{Failure:}\\ Model outputs ineffective chain of thoughts, and giving the wrong answer in the wrong format(ground truth is B))
			
			\vspace{0.2cm}
			\textbf{User:}\\ Question: What is the sum of the index of top 30 smallest numbers in the list? Options: A: 1566, B: 1572, C: 1575, D: 1573, E: 1577, F: 1564, G: 1569, H: 1567 Data: [-2568.1542995196, 1035.3645340299...(100 numbers)]
			
			\vspace{0.3cm}
			\textbf{Metamath-Llemma-7B Model:} \\ The data you receive is a list of numbers. To find the sum of the index of the top 30 smallest numbers, we need to find the 30 smallest numbers in the list and then calculate their sum. To find the 30 smallest numbers, we can use the min function in Python. The min function takes a list as input and returns the smallest number in the list. We can use the min function to find the 30 smallest numbers in the list. Now, we need to calculate the sum of these 30 smallest numbers. To do this, we can use the sum function in Python. The sum function takes a list as input and returns the sum of all the numbers in the list. We can use the sum function to calculate the sum of the 30 smallest numbers. Finally, we need to find the index of the sum. To do this, we can use the index function in Python. The index function takes a number as input and returns the index of that number in the list. We can use the index function to find the index of the sum. Therefore, the answer is (correct option). \#\#\#\# 1566 The answer is: 1566
		\end{examplebox}
		\caption{Failure type and conversation record of Metamath-Llemma-7B. The model attempts to use programming language to solve the problem, but it does not provide executable lines of code, only providing vague description of certain functions instead. After outputting the vague solution, the model outputs an answer, which is different from the ground truth and not in the expected format(We expect the model to provide the answer with a capital letter corresponding to the correct option).}
		\label{fig:fail_llemma}
	\end{figure*}

	\begin{figure*}
	\begin{examplebox}{MAmmoTH-13B}
		\textbf{Failure:}\\ Model outputs ineffective chain of thoughts, and giving the wrong answer in the wrong format(ground truth is B))
		
		\vspace{0.2cm}
		\textbf{User:}\\ Question: What is the sum of the index of top 30 smallest numbers in the list? Options: A: 1566, B: 1572, C: 1575, D: 1573, E: 1577, F: 1564, G: 1569, H: 1567 Data: [-2568.1542995196, 1035.3645340299...(100 numbers)]
		
		\vspace{0.3cm}
		\textbf{MAmmoTH-13B Model:}\\  Notice that there are two numbers in your list that have the smallest index: 4346.5174909575 and 8169.6394989672. They have the smallest index of 4346 and 8169 respectively. Since in the options the first number is the index of the smallest number, then the correct option is D.\\             Option: D
	\end{examplebox}
	\caption{Failure type and conversation record of Mammoth-13B. The model fail to understand the problem's instruction, and does not provide reasonable output. Moreover, it provides the wrong answer in incorrect format, just like Metamath-Llemma-7B(We expect the model to provide the answer with a capital letter corresponding to the correct option).}
	\label{fig:fail_mammoth}
	\end{figure*}

\end{document}